%% file: root.tex
\documentclass[letterpaper, 10 pt, conference]{ieeeconf}  

\IEEEoverridecommandlockouts                              

\usepackage{amsmath}
\usepackage{amssymb}
\usepackage{amsfonts}
\usepackage{bm}
\usepackage{adjustbox}
\usepackage{multirow}
\usepackage{wrapfig}
\usepackage[usenames,dvipsnames]{xcolor}
\usepackage{algorithm}
\usepackage{algpseudocode}
\usepackage{caption}
\usepackage{subcaption}
\usepackage{mathtools}
\usepackage{url}
\usepackage{hyperref}

\usepackage{booktabs}       
\usepackage{subcaption}
\usepackage{microtype}      
\usepackage{pifont}

\usepackage{nicefrac}

\title{

\input{includes/title}}

\author{

\input{includes/authors.tex}}

\begin{document}

\maketitle
\thispagestyle{empty}
\pagestyle{empty}

\begin{abstract}
\input{includes/abstract}

\end{abstract}

\vspace{-1mm}
\section{Introduction}
\vspace{-1mm}
\label{sec:introduction}

Benchmarking has been crucial for successfully boosting the rate of research in machine learning-related communities. With a common benchmark for experiments, different algorithms can be compared in a fair and standardized manner, and a relative improvement between them can be established. A well-known benchmark in AI research is the ImageNet dataset~\cite{deng2009imagenet} and its associated challenge~\cite{russakovsky2015imagenet}. It has been used to train and test image classification methods and can be attributed to the rise of network architecture revolution in deep learning. Similarly, the KITTI dataset~\cite{geiger2013vision} has been widely used to study problems related to autonomous driving such as car detection and tracking, lane segmentation, and odometry estimation. In NLP too, a wide variety of benchmarks were introduced with different levels of difficulty, such as GLUE~\cite{wang2018glue}, SuperGLUE~\cite{wang2019superglue} and MMLU~\cite{hendrycks2020measuring}. Recently, researchers have proposed several benchmarks for different robotics tasks, such as the NIST assembly boards~\cite{kimble2020benchmarking} for assembly and the GRASPA~\cite{bottarel2020graspa} for grasping. However, most of them have not seen widespread adoption and usage, thereby limiting the rate of research progress as it becomes harder to compare methods and investigate their true applicability to a problem. 

The key difficulty in robotics benchmarking is that robot tasks in the real world involve a complex pipeline compared to running experiments on fixed test datasets. 
Real-world tasks don't usually have a common test dataset, and it is difficult to ensure replicability of the test scenarios across different research labs. Despite this, the robotics community has made progress in several interesting directions for benchmarking, especially robot manipulation. First, for assembly tasks, the NIST assembly boards~\cite{kimble2020benchmarking} provide a reproducible test bed. Second, the YCB Object and Model Set~\cite{calli2015ycb} provides a way for easily available objects with their 3D mesh models. Although these objects are limited to dozens of rigid objects, they are useful for testing manipulation in an accessible manner. Third, several benchmarking protocols for real-world manipulation are introduced using YCB objects. These protocols show how to arrange objects in the real world to create reproducible scenes. For example, \cite{bekiroglu2019benchmarking} introduces a grasp planning protocol that uses drawing on the robot workspace for the placement of objects. GRASPA~\cite{bottarel2020graspa} uses printable boards with AR tags for reproducible placement of YCB objects. However, \cite{bekiroglu2019benchmarking} cannot guarantee accurate reproducible object placement, while \cite{bottarel2020graspa} has to use AR tags which create unnatural scenes for testing.

In this work, we introduce a new benchmark, SceneReplica, for real-world robot manipulation to overcome the limitations of existing benchmarks. We use 16 YCB objects which can be easily purchased either online or in-person. Second, we created a dataset of 20 scenes with 5 YCB objects in each scene as a test bed for pick-and-place. 
We utilize a procedural scene generation for dataset design~\cite{newbury2023deep} in simulation to create these 20 scenes. The design considers the reachability space of a robot and the diversity of poses of objects in building the real-world test bed. Finally, we provide an easy-to-use tool to guide users in reproducing these 20 scenes in the real world without using any AR tag or external equipment. To the best of our knowledge, SceneReplica is the only benchmark for real-world reproducible scenes without using AR markers or external cutouts to guide object placement.

\begin{table*}
\centering
\begin{adjustbox}{max width=0.9\textwidth}
\begin{tabular}{@{}lccccc@{}}

\textbf{Benchmark}            & \textbf{Type} & \textbf{Task} & \textbf{Objects} &  \textbf{AR Tag-Free} & \textbf{Scene Reproducibility} \\ \midrule

Meta-World~\cite{yu2020meta} & Simulation & 50 tasks & Synthetic & \textcolor{ForestGreen}{\ding{51}} & \textcolor{ForestGreen}{\ding{51}} \\

RLBench~\cite{james2020rlbench} & Simulation & 100 Tasks & Synthetic & \textcolor{ForestGreen}{\ding{51}} & \textcolor{ForestGreen}{\ding{51}}\\

robosuite~\cite{zhu2020robosuite} & Simulation & 9 Tasks & Synthetic & \textcolor{ForestGreen}{\ding{51}} & \textcolor{ForestGreen}{\ding{51}} \\

\hline

Grasp Planning Protocol~\cite{bekiroglu2019benchmarking} & Real & Grasp Planning & YCB (single) & \textcolor{ForestGreen}{\ding{51}} & \textcolor{red}{\ding{55}}  \\

NIST Assembly~\cite{kimble2020benchmarking}           & Real                  & Assembly           & Task Boards        & \textcolor{ForestGreen}{\ding{51}}  & \textcolor{ForestGreen}{\ding{51}}           \\

FurnitureBench~\cite{heo2023furniturebench} & Real & Assembly & 3D Printing & \textcolor{red}{\ding{55}} & \textcolor{ForestGreen}{\ding{51}}  \\

GRASPA~\cite{bottarel2020graspa} & Real & Grasping & YCB (clutter) & \textcolor{red}{\ding{55}} & \textcolor{ForestGreen}{\ding{51}} \\

OCRTOC~\cite{liu2021ocrtoc} & Real & Rearrangement & YCB + Others & \textcolor{ForestGreen}{\ding{51}} & \textcolor{red}{\ding{55}} \\

RB2~\cite{dasari2022rb2} & Real &  Pouring, Scooping, Zipping, Insertion & Others & \textcolor{ForestGreen}{\ding{51}} & \textcolor{red}{\ding{55}} \\

Box and Blocks Test~\cite{morgan2019benchmarking} & Real & Pick-and-Place & Blocks & \textcolor{ForestGreen}{\ding{51}} & \textcolor{red}{\ding{55}}  \\

\textbf{SceneReplica (Ours)} & Real & Pick-and-Place & YCB (clutter) & \textcolor{ForestGreen}{\ding{51}} & \textcolor{ForestGreen}{\ding{51}} \\

\bottomrule
\end{tabular}
\end{adjustbox}
\caption{Comparison between different robotic manipulation benchmarks.}
\label{tab:benchmark}
\vspace{-2mm}
\end{table*}


Another contribution of our work is the benchmarking of two main paradigms for 6D robotic grasping using SceneReplica: model-based grasping and model-free grasping. Model-based grasping leverages 3D models of objects for perception and planning. It consists of a pipeline using 6D object pose estimation for perception~\cite{xiang2017posecnn,tremblay2018deep,deng2021poserbpf}, 3D model-based grasp planning~\cite{miller2004graspit,diankov2008openrave} and motion planning and motion control~\cite{sucan2012open} to reach the planned grasps. Model-free grasping does not assume the availability of 3D object models. It consists of a pipeline of unseen object instance segmentation~\cite{xie2020best,xiang2021learning,lu2022mean}, object point cloud-based grasp planning~\cite{mousavian20196,sundermeyer2021contact} and motion planning and motion control~\cite{sucan2012open} to reach the planned grasps. Different object perception, grasping planning, motion planning, and control methods can be evaluated within the two 6D robotic grasping paradigms. This high-level view of robot manipulation benchmarking is a departure from existing benchmarks which are specific to some tasks or do not give details about the complete pipeline. We also perform experiments and provide our analyses on a selected set of representative methods. To the best of our knowledge, it is the first effort in which model-based and model-free grasping pipelines are systematically evaluated and fairly compared to each other using the same real-world testing scenes.

By creating SceneReplica, we provide the robotics community with an easy-to-setup, reproducible real-world benchmark for robot manipulation. The core method for SceneReplica involving scene generation in simulation and replication in real-world with rendered images is applicable to \textit{any robotic platform}, which allows for testing with different robots. We plan to maintain the benchmark continuously and evaluate additional and upcoming perception, planning, and control methods on it. It is our hope that these results, which combine effects from the entire pipeline, can serve as a reference for robot manipulation research.


\vspace{-1mm}
\section{Related Work}
\vspace{-1mm}
\label{sec:related-work}


\textbf{Simulation-based Robotic Benchmarks.} Since reproducibility in the real world is challenging, the setup of testing environments in physics simulators provides a workaround for benchmarking in robotics. For example, the Meta-World~\cite{yu2020meta} benchmark introduces 50 robotic manipulation tasks, which can be used to study meta-reinforcement learning and multitask learning. The robosuite~\cite{zhu2020robosuite} based on the MuJoCo physics engine~\cite{todorov2012mujoco} supports 9 manipulation tasks for different robots. The RLBench benchmark~\cite{james2020rlbench} increases the number of tasks to 100. The advantage of using simulation is that reproducibility is guaranteed for evaluation, and a large number of tasks can be used. However, due to the sim-to-real gap, the performance in simulation cannot simply be transferred to the real world. Although there is concurrent research effort on improving the physics simulators, nuances in the real world such as lighting, friction, external perturbation, etc., are extremely difficult to model in simulation. This motivates our need to have real-world benchmarks for robotics.

\textbf{Real-world Robotic Benchmarks.} We summarize a list of representative efforts to benchmark robot manipulation in the real world in Table~\ref{tab:benchmark}. According to the evaluation tasks, they can be classified into assembly~\cite{kimble2020benchmarking,heo2023furniturebench}, grasping~\cite{bekiroglu2019benchmarking,morgan2019benchmarking,bottarel2020graspa, rudorfer2022burg-toolkit}, rearrangement~\cite{liu2021ocrtoc}, and others (pouring, scooping zipping, insertion in~\cite{dasari2022rb2}). In terms of scene reproducibility in evaluation, NIST Assembly~\cite{kimble2020benchmarking}, FurnitureBench~\cite{heo2023furniturebench} and GRASPA~\cite{bottarel2020graspa} are the only three benchmarks whose test scenes are reproducible. NIST Assembly uses a fixed set of assembly task boards, while FurnitureBench and GRASPA use AR markers to guide the scene creation process. For grasping and pick-and-place, SceneReplica is the only benchmark that can create locally reproducible scenes without using AR markers. Another flavor of benchmarks includes TOTO~\cite{zhou2023train-toto} and RRC~\cite{bauer2022real-rrc} which provide remote access to shared hardware, to benchmark common tasks. SceneReplica can be used in conjunction with such benchmarks in a ranking system as proposed by RB2~\cite{dasari2022rb2} where researchers have the option of local and remote evaluation of their proposed methods.


\vspace{-1mm}
\section{The SceneReplica Benchmark}
\vspace{-1mm}
\label{sec:method}

Our goal is to introduce reproducible scenes of objects for robot manipulation, where researchers can set up the same set of scenes in different environments for evaluation. Currently, we focus on tabletop cluttered scenes for pick-and-place where the clutter is in terms of a packed scene with some level of separation between objects. This is in contrast with other classes of cluttered scenes where objects are piled on top of each other, making it harder to replicate exactly in the real-world or simulation. Overall, the scene generation pipeline happens in Gazebo with access to ground truth object data for constraint checking over the scenes.

\begin{figure}
    \centering
\includegraphics[width=0.85\columnwidth]{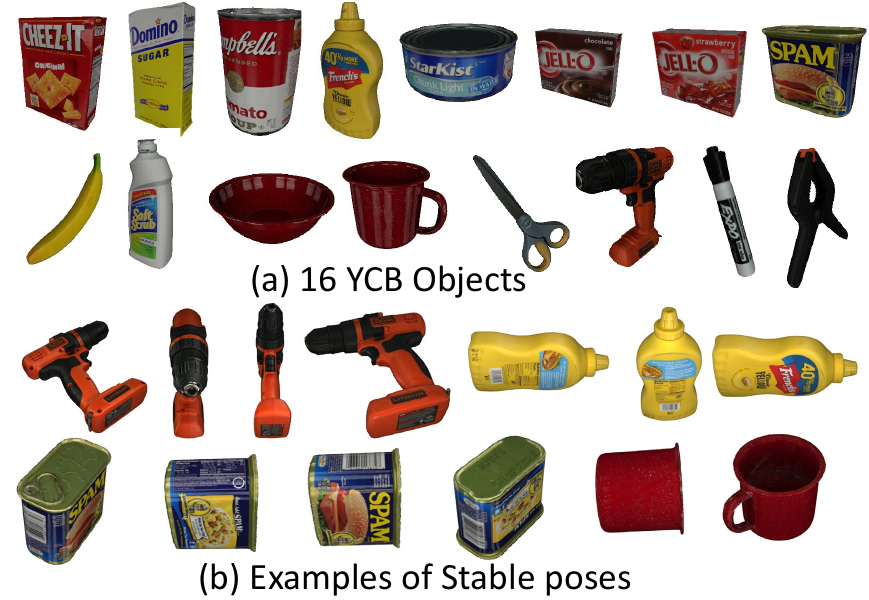}
\vspace{-2mm}
    \caption{16 YCB objects are used in SceneReplica.}
    \label{fig:ycb}
    \vspace{-2mm}
\end{figure}

\subsection{Creating Scenes in Simulation}

\textbf{Object Set.} 16 YCB objects~\cite{calli2015ycb} are used in our benchmark to create scenes. Their 3D mesh models are shown in Fig.~\ref{fig:ycb}. We select these objects because i) they can be easily purchased online or in grocery stores; ii) they are a subset of the YCB Video dataset~\cite{xiang2017posecnn} which has been widely used to study 6D object pose estimation; (iii) they can reliably fit within a parallel-jaw gripper. Therefore, we can leverage existing 6D object pose estimation methods for experiments, for example,~\cite{xiang2017posecnn,deng2021poserbpf,Wang_2021_GDRN}. 

\textbf{Stable Poses of Objects.} We create scenes in a Gazebo simulation, where stable poses of objects are used. We do not create random clutter via a piling procedure for object spawning as it requires careful tuning and is not easily reproducible in real world. 
Therefore, we opt to compute stable poses of objects, which are the object orientations when they rest statically on a planar surface. Such stable poses are also safe to use in real world scenes as the chance of an object falling over and disturbing the scene is low. We used the compute\_stable\_poses() function in trimesh to obtain such poses, and Fig.~\ref{fig:ycb} shows examples of stable poses of some YCB objects.

\begin{figure}
    \centering
\includegraphics[width=0.85\columnwidth]{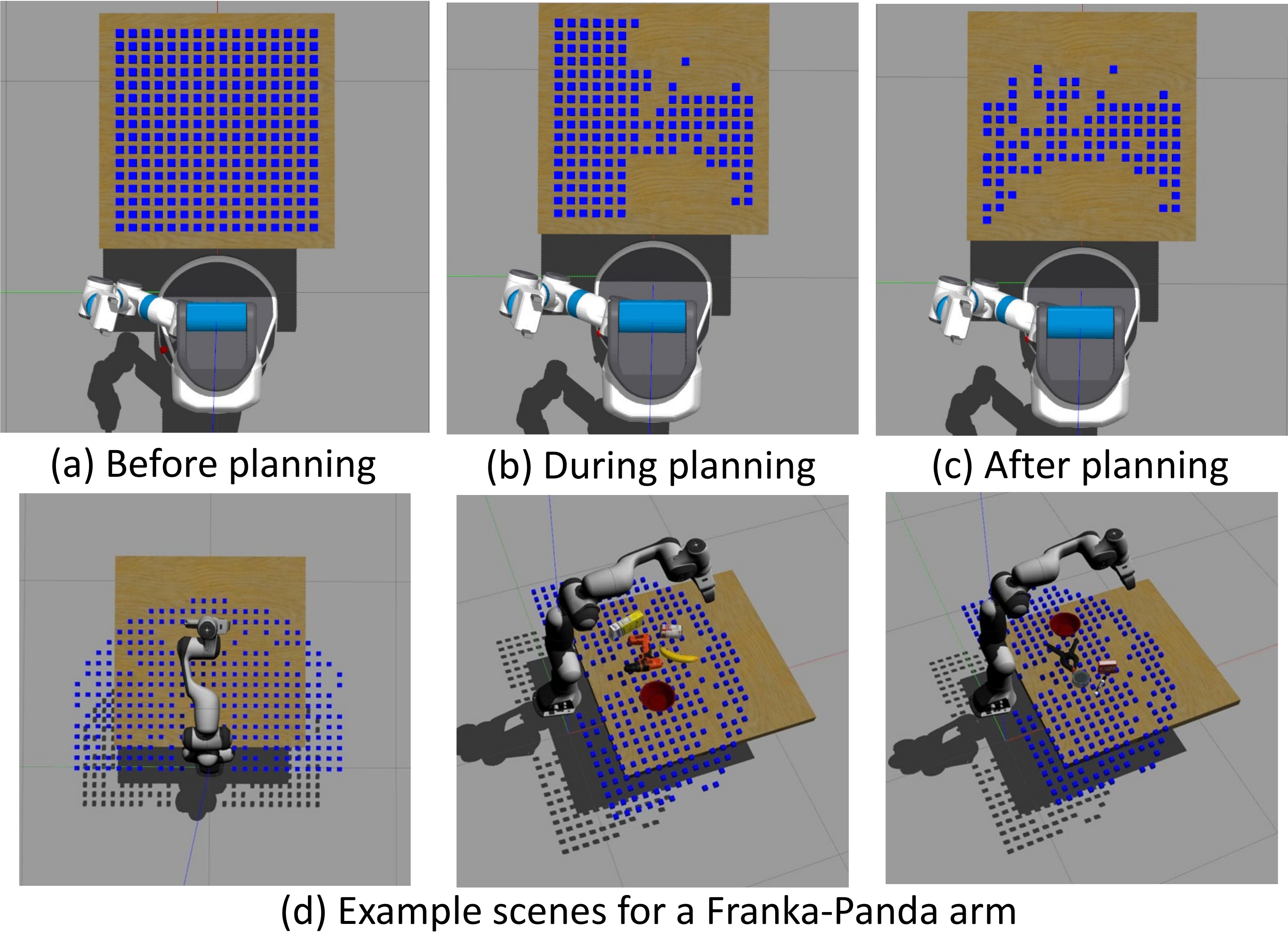}
\vspace{-2mm}
    \caption{(a,b) Illustration of motion planning check to filter out reachable locations. (c) Blue cubes remaining on table after planning indicate reachable locations. (d) The algorithm can be extended to different robots.}
    \label{fig:reach}
    \vspace{-2mm}
\end{figure}

\textbf{Determining the Reachable Space of a Robot.} When placing these objects on a table, we need to place them within the reach of the robot to avoid motion planning failures. We employ a simple algorithm as depicted in Fig.~\ref{fig:reach} to compute the reachable regions of a tabletop. In Step I, the robot model is loaded in Gazebo with its arm stowed. Next, a table of dimension $1m$ x $1m$ x $0.745m$ is placed at an offset of $(0.8,0,0)$ with respect to the robot model, where the $x$-axis is the forward direction of the robot. In Step II, the surface of the table is partitioned into grids $16\times 16$ and a block of size $3cm \times 3cm$ is placed in the center of each cell of the grid. In Step III, we determine whether there is a feasible motion plan for the stand-off pose to each cube. Blocks that do not have a feasible plan will be removed from the scene and earlier steps are repeated. The blocks surviving the removal denote reachable locations which are then used to spawn objects for scene creation.

\begin{figure*}
    \centering
\includegraphics[width=1.8\columnwidth]{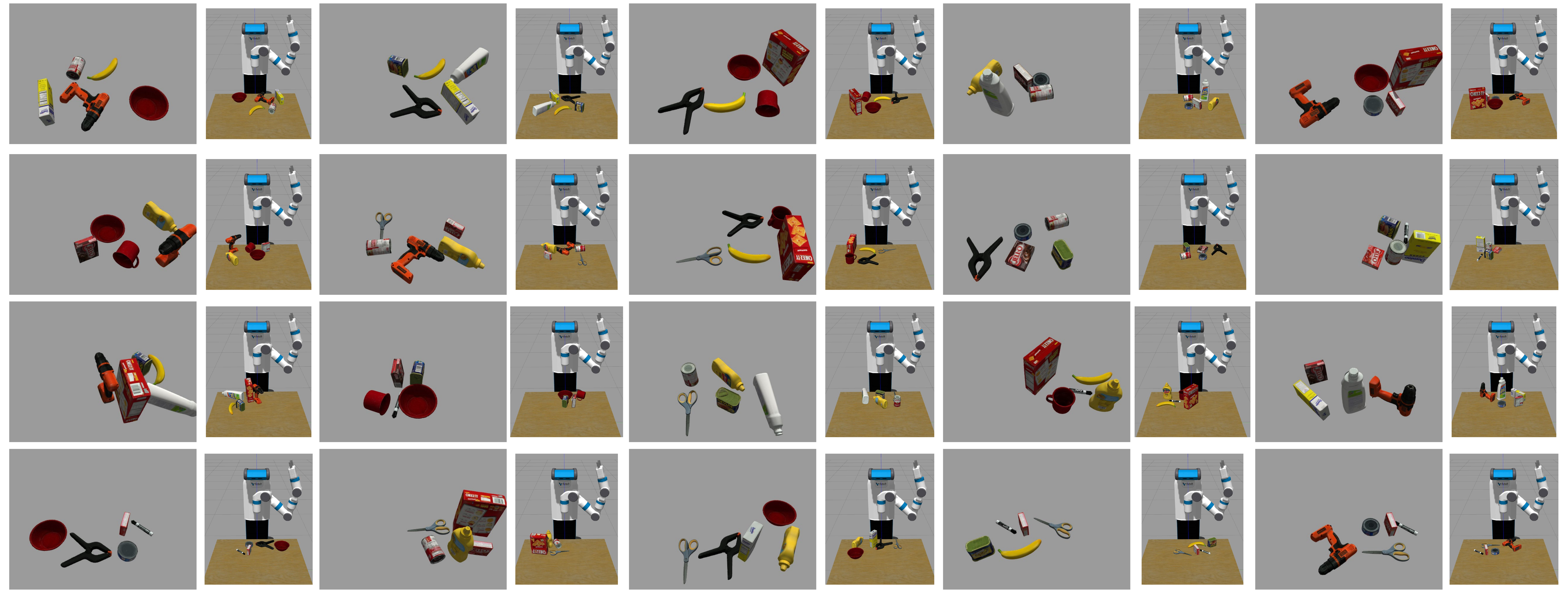}
    \caption{20 scenes in our SceneReplica benchmark with 5 YCB objects in each scene.}
    \label{fig:scenes}
    \vspace{-2mm}
\end{figure*}

\textbf{Scene Generation.} Using the stable poses of the 16 YCB objects, we can generate random scenes in Gazebo by placing them in the reachable locations of the Fetch robot. We limit the maximum number of objects in a scene to five and sequentially add objects to the scene one by one. Initially, an object is placed randomly on the table. When adding the next object, we randomly sample a nearby reachable location with an existing object that is collision-free between these objects. Stable poses of objects are also randomly sampled during placement. This process repeats until five objects are added to the scene. By sampling nearby reachable locations, we tend to generate cluttered scenes where objects occlude each other, thereby creating a strong benchmark. 

\textbf{Scene Selection.} After generating a set of scenes, we target to select a set of 20 scenes with the following properties. i) object count distribution across scenes should be balanced. We define a range $[C_\text{min}, C_\text{max}]$ for the object counts. ii) The selected scenes should adequately cover the stable poses of each object. iii) All the objects in a selected scene should be reachable from the robot.  To check the reachability of an objects in a scene, we check for existence of valid motion plans to (offline-generated) grasps around the object.

With these three properties, our algorithm for scene selection works as follows. We generate a large number of scene sets. Each scene set contains 20 scenes, and it satisfies properties i) and iii). In addition, each set of scenes is given a score on how many stable poses of these 16 YCB objects are covered by the 20 scenes in it. This acts as a measure of the diversity of a set. Finally, we select the set of scenes with the highest pose coverage score. Fig.~\ref{fig:scenes} shows the final 20 scenes selected for SceneReplica.

\textbf{Transfer to other robots.} Generating in simulation enables access to complete state information, allowing for a seamless sim2real transfer for reproducible real-world experiments across different robotic platforms (e.g.~\ref{fig:reach}) For use with a new robot, users just need to re-render the reference images (e.g.~\ref{fig:scenes}) for the scenes in simulation via the robot camera's parameters. Users then recreate the scenes in real world using the rendered images which act as the new reference images. Further details are presented in the supplementary material.

\begin{figure*}
    \centering
\includegraphics[width=1.8\columnwidth]{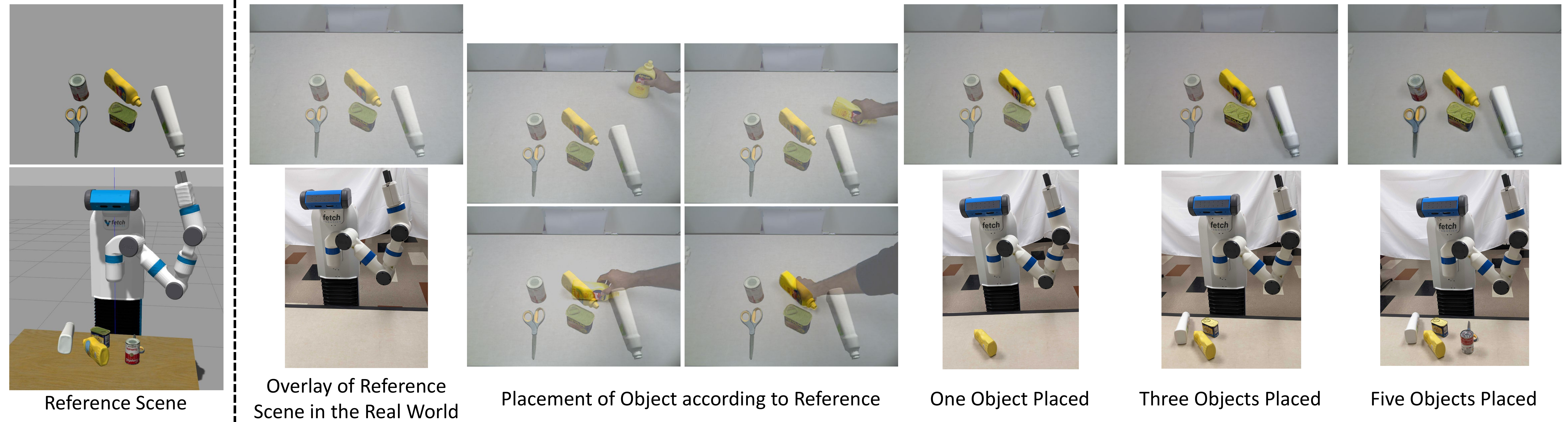}
\vspace{-1mm}
    \caption{The process of replicating a scene in the real world. The image of the reference scene is overlaid on the image of the real camera to guide how to place objects into the real-world scene.}
    \label{fig:replication}
    \vspace{-3mm}
\end{figure*}

\subsection{Reproducing Scenes in the Real World}
\vspace{-1mm}

To use the 20 selected scenes for experiments, we need to replicate these scenes in the real world. Previous benchmarks rely on AR markers to create reproducible scenes~\cite{bottarel2020graspa,heo2023furniturebench}. In contrast, we show that directly using the scene images in Fig.~\ref{fig:scenes} as reference images, we can place objects in the real world as in the simulator. The process of replicating a scene in the real world is illustrated in Fig.~\ref{fig:replication}. 
First, the Fetch camera pose in the real world is set to be the same as in the simulation. Therefore, the camera extrinsic, that is, the SE(3) transformation from the robot base frame to the RGB camera frame, is always the same. This can be done by controlling the Fetch head pose. Second, maintain uniform table height at $0.745m$ for both real-world and simulated environments, achievable with a height-adjustable table. Third, with the same camera extrinsic and table height, we can overlay the half-transparent version of an RGB image of a reference scene to the real RGB image from the Fetch camera for object placement guidance. Finally, the user can place objects by interactively checking the overlap between the reference image and the real-world image for accurate alignment (Fig.~\ref{fig:replication}). 

\begin{wrapfigure}{r}{0.25\textwidth}
    \centering
    \vspace{-2mm}
    \includegraphics[width=0.25\textwidth]{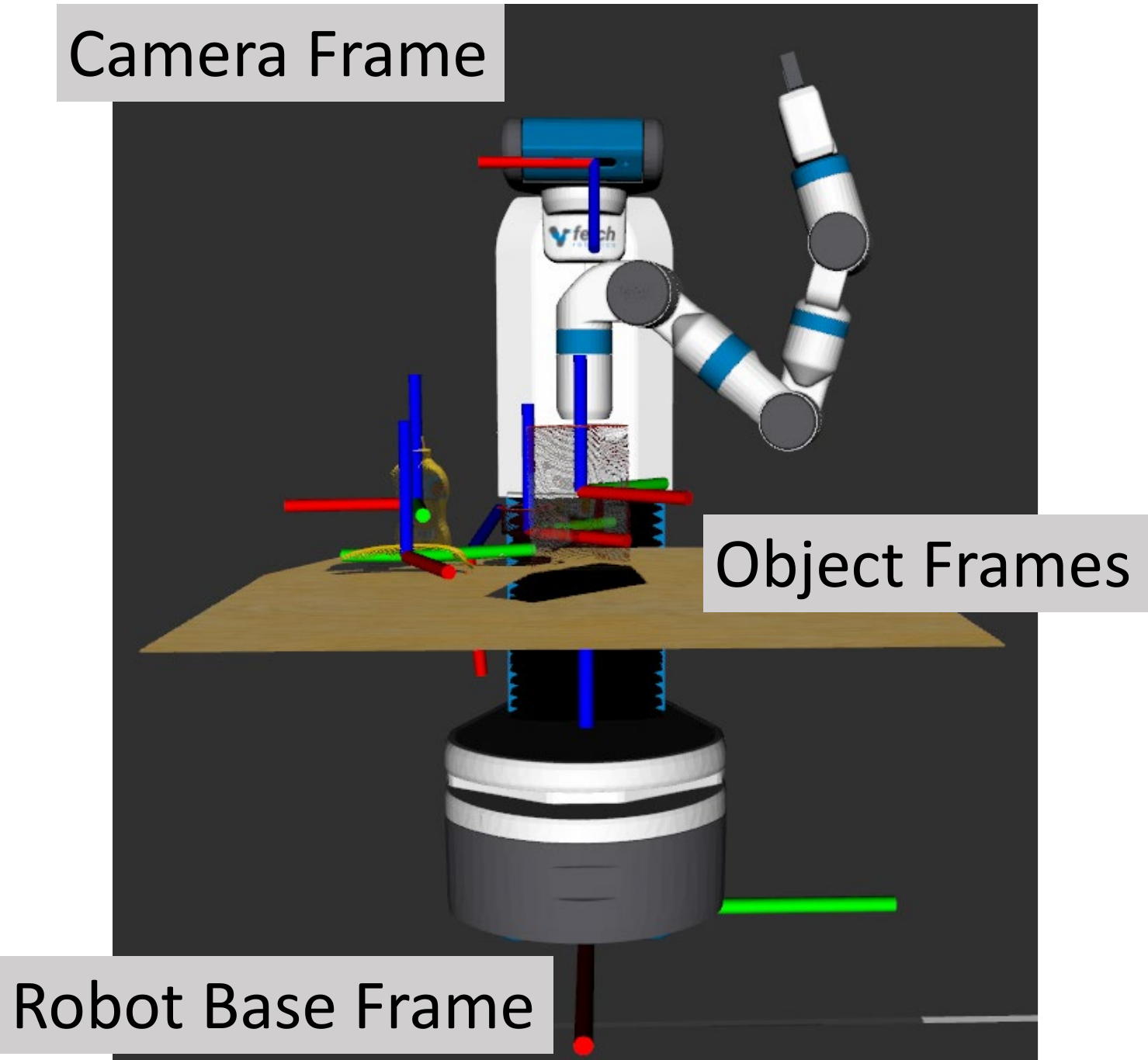}
    \caption{Illustration of the coordinate frames.}
    \vspace{-3.5mm}
    \label{fig:frames}
\end{wrapfigure}


By matching camera extrinsic between simulation and reality, object poses in the robot base frame replicate those in the camera frame for manipulation. This procedure is transferable to different robot platforms using SceneReplica, as only the reference images need to be re-generated by the new camera.

Fig.~\ref{fig:frames} illustrates these coordinate frames used for our experiments.

\section{Model-based and Model-free 6D Grasping}
\vspace{-1mm}
\label{sec:dataset}


In the SceneReplica benchmark, we assess tabletop pick-and-place tasks involving grasping, lifting, gripper rotation, and object drop-off, in sequence. Metrics are collected for each stage, with success categorized as (1) grasping and (2) pick-place success. Grasping-success indicates successful grasp but failure in later steps, while pick-and-place success denotes completion of the entire pipeline. This task necessitates 6D grasping, controlling both gripper orientation and translation, distinct from top-down grasping. 



\vspace{-2mm}
\subsection{Model-based 6D Grasping} 
\begin{figure*}[h]
    \centering
    \includegraphics[width=1.8\columnwidth]{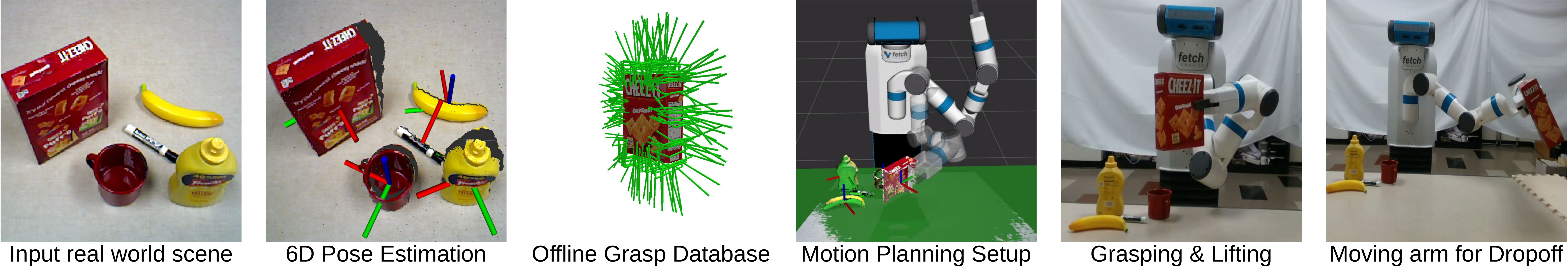}
    \vspace{-2mm}
    \caption{Model-based grasping pipeline used for benchmarking}
    \label{fig:pipeline-modelbased}
    \vspace{-3mm}
\end{figure*}

\begin{figure*}[h]
    \centering
    \includegraphics[width=1.8\columnwidth]{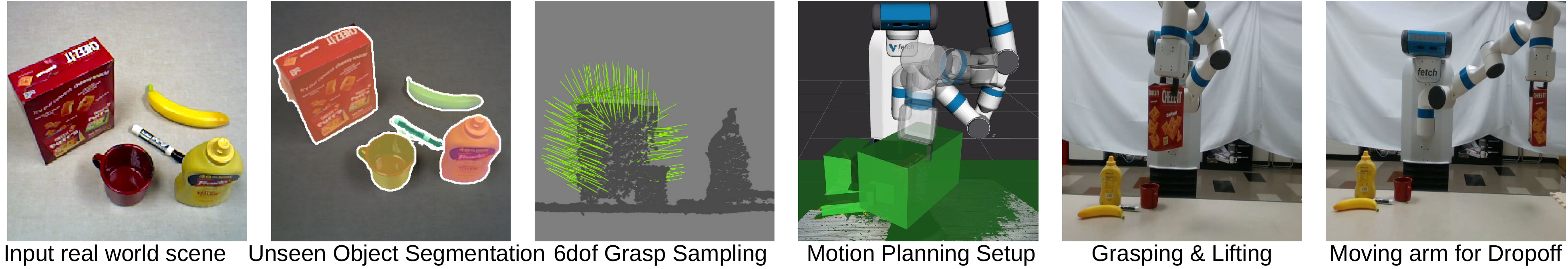}
    \vspace{-1mm}
    \caption{Model-free grasping pipeline used for benchmarking}
    \label{fig:pipeline-modelfree}
    \vspace{-7mm}
\end{figure*}



\textbf{Object Perception.} When 3D models of objects are available, we deal with 6D object pose estimation, an active research area in robotics and computer vision \cite{workshop}. It, determines 3D orientation and translation with respect to the camera. Though pose accuracy can be assessed on datasets, its impact on grasping remains unclear. SceneReplica integrates pose estimation with planning and control for 6D grasping, enabling analysis of pose accuracy's effect on success rates

\textbf{Grasp Planning.} Model-based grasp planning methods such as GraspIt!~\cite{miller2004graspit} and OpenRAVE~\cite{diankov2008openrave} use 3D models of objects in physics simulators to generate grasps. We use GraspIt! in our experiments. We initially generated numerous grasps for each YCB object in GraspIt! then sampled a fixed amount using farthest point sampling for grasp pose diversity.

\textbf{Model-based Motion Planning.} After estimating the 6D pose, planned grasps on the object model are transformed to the robot base frame. Each grasp is checked for a feasible trajectory to reach it using motion planning algorithms from OMPL~\cite{sucan2012open}. Upon obtaining a trajectory, it is executed to grasp the target. All Object pose estimation enables utilization of object 3D models as obstacles in the planning scene.

\subsection{Model-free 6D Grasping}
\vspace{-1mm}


\textbf{Object Perception.} When there is no 3D model of objects available, we can tackle the \emph{unseen object instance segmentation (UOIS)} problem~\cite{xie2020best,xiang2021learning,lu2022mean} to segment arbitrary objects. Once the objects are segmented, model-free grasp planning methods can be used to generate grasps of the segmented objects. UOIS methods are usually trained using synthetic images of a large number of objects. They learn the concepts of objects and are able to segment novel objects in the real world.

\textbf{Grasp Planning.} Grasp planning methods such as 6D GraspNet~\cite{mousavian20196,sundermeyer2021contact} take point clouds of objects or whole scenes as input and output grasps of objects in the scenes. Neural networks trained for grasp synthesis, like those for unseen object instance segmentation, learn from large datasets of object-grasp pairs, allowing them to plan grasps for new objects using partial point clouds.


\textbf{Motion Planning.} Once the target object is segmented and grasps are synthesized, a motion planning algorithm generates a trajectory towards a planned grasp. Instead of using 3D object models, oriented 3D bounding boxes for the segmented objects are computed for obstacle avoidance during planning.


\section{Experiments}
\label{sec:experiments}


In pick-and-place tasks, the primary evaluation metric is the success rate of transporting objects between the locations. This success relies on the performance of all the framework components. Even with accurate perception, errors in grasp planning can prevent the robot from picking up objects effectively. Therefore, we evaluated the success rate, perception evaluation and failure analysis of pick-and-place of 100 trials [20 scenes $\times$ 5 objects per scene]  across diverse grasping frameworks presented in Table~\ref{tab:grasp_methods}. 

\textbf{Failure Analysis.} Any error during the experiment is classified into 3 categories: (i) Perception error (ii) Planning error (iii) Execution error. We use the average distance for symmetric objects (ADD-S) metric mentioned in~\cite{xiang2017posecnn} to evaluate 6D object pose estimation. For unseen object instance segmentation, we follow~\cite{xie2020best,xiang2021learning} to compute precision, recall, F-measure for region \& boundary overlaps.

(i) A \textit{Perception Error} occurs when grasping fails due to either pose estimation (in model-based grasping) or segmentation (in model-free grasping) inaccuracies. We denote a pose estimation error when the grasp fails and ADD-S $> 0.1 * d$ (d = 3D object model's diameter). For segmentation methods, we consider the prediction a failure if the overlap F-measure is $< 75$\%. (ii) Despite no obvious perception error, if the manipulator fails to grasp the object, it indicates issues with either the grasp strategy or trajectory planning. In such a scenario we record a \textit{Planning Error} for the experiment. (iii) When the robot successfully grasps the object but fails to drop it off due to slipping or any other reason, it is classified as an \textit{Execution Error}.

\begin{table*}
\vspace{-1mm}
\centering
\resizebox{0.9\textwidth}{!}{
\begin{tabular}{c|c|c|c|c|ccc}
\hline
Method \# & Perception & Grasp Planning & Motion Planning & Control & Ordering & Pick-and-Place Success & Grasping Success  \\ \hline

\multicolumn{8}{c}{Model-based Grasping} \\ \hline
1 & PoseRBPF~\cite{deng2021poserbpf} & GraspIt!~\cite{miller2004graspit} + Top-down & OMPL~\cite{sucan2012open} & 
 MoveIt  & Near-to-far & 58 / 100 & 64 / 100\\

1 & PoseRBPF~\cite{deng2021poserbpf} & GraspIt!~\cite{miller2004graspit} + Top-down & OMPL~\cite{sucan2012open} & MoveIt  & Fixed & 59 / 100 & 59 / 100\\

2 & PoseCNN~\cite{xiang2017posecnn} & GraspIt!~\cite{miller2004graspit} + Top-down & OMPL~\cite{sucan2012open} & 
 MoveIt  & Near-to-far & 47 / 100 & 48 / 100\\

2 & PoseCNN~\cite{xiang2017posecnn}  & GraspIt!~\cite{miller2004graspit} + Top-down & OMPL~\cite{sucan2012open} & MoveIt  & Fixed & 40 / 100 & 45 / 100 \\

3 & GDRNPP~\cite{Wang_2021_GDRN,liu2022gdrnpp_bop} & GraspIt!~\cite{miller2004graspit} + Top-down & OMPL~\cite{sucan2012open} & 
 MoveIt  & Near-to-far & \textbf{66 / 100} & 69 / 100\\

3 & GDRNPP~\cite{Wang_2021_GDRN,liu2022gdrnpp_bop} & GraspIt!~\cite{miller2004graspit} + Top-down & OMPL~\cite{sucan2012open} & MoveIt  & Fixed & 62 / 100 & 64 / 100\\ \hline

\multicolumn{8}{c}{Model-free Grasping} \\ \hline

4 & UCN~\cite{xiang2021learning} & GraspNet~\cite{mousavian20196} + Top-down & OMPL~\cite{sucan2012open} & 
 MoveIt  & Near-to-far & 43 / 100  & 46 / 100\\

 4 & UCN~\cite{xiang2021learning} & GraspNet~\cite{mousavian20196} + Top-down & OMPL~\cite{sucan2012open} & 
 MoveIt  & Fixed & 37 / 100 & 40 / 100 \\

 5 & UCN~\cite{xiang2021learning} & Contact-graspnet~\cite{sundermeyer2021contact} + Top-down & OMPL~\cite{sucan2012open} & 
 MoveIt & Near-to-far & 60 / 100  & 63 / 100\\

  5 & UCN~\cite{xiang2021learning} & Contact-graspnet~\cite{sundermeyer2021contact} + Top-down & OMPL~\cite{sucan2012open} & 
 MoveIt  & Fixed & 60 / 100 & 64 / 100 \\

 6 & MSMFormer~\cite{lu2022mean} & GraspNet~\cite{mousavian20196} + Top-down & OMPL~\cite{sucan2012open} & 
 MoveIt  & Near-to-far & 38 / 100 & 41 / 100 \\

 6 & MSMFormer~\cite{lu2022mean} & GraspNet~\cite{mousavian20196} + Top-down & OMPL~\cite{sucan2012open} & 
 MoveIt  & Fixed & 36 / 100 & 41 / 100 \\

 7 &  MSMFormer~\cite{lu2022mean} &  Contact-graspnet~\cite{sundermeyer2021contact} + Top-down & OMPL~\cite{sucan2012open} & 
 MoveIt  & Near-to-far & 57 / 100  & 65 / 100\\ 

 7 &  MSMFormer~\cite{lu2022mean} &  Contact-graspnet~\cite{sundermeyer2021contact} + Top-down & OMPL~\cite{sucan2012open} & 
 MoveIt  & Fixed & 61 / 100  & \textbf{70 / 100}\\

8 & MSMFormer~\cite{lu2022mean} & Top-down & OMPL~\cite{sucan2012open} & 
 MoveIt  & Fixed & 56 / 100  & 59 / 100\\
\hline
\multicolumn{8}{c}{End-to-end Learning-based Grasping} \\ \hline

9 & \multicolumn{2}{c|}{Dex-Net 2.0~\cite{mahler2017dex} (Top-Down Grasping)} & OMPL~\cite{sucan2012open} &  MoveIt & Algorithmic &  43 /100   &  51 / 100 \\

\hline

\multicolumn{8}{c}{Ground truth pose-based Grasping} \\ \hline
10 &Ground truth object pose & GraspIt!~\cite{miller2004graspit} + Top-down & OMPL~\cite{sucan2012open} & 
 MoveIt  & Near-to-far & 78 / 100 & 82 / 100\\ 
 11 &Ground truth object pose & GraspIt!~\cite{miller2004graspit} + Top-down & OMPL~\cite{sucan2012open} & 
 MoveIt  & Fixed & 78 / 100 & 87 / 100\\ 
 \hline
\end{tabular}
}
\vspace{-1mm}
\caption{Different grasping frameworks evaluated on SceneReplica using a Fetch mobile manipulator.}
\label{tab:grasp_methods}
\end{table*}

\begin{table}
  \centering
\scalebox{0.65}{
\begin{tabular}{|c|c||c|ccccccc|}
\hline
Pose Method & ADD-S &

\multirow{2}{*}{Segmentation Method}                         & \multicolumn{3}{c|}{Overlap}   & \multicolumn{3}{c|}{Boundary}  &      \\  \cline{1-2}

PoseCNN~\cite{xiang2017posecnn} &  66.01 &       & P & R & \multicolumn{1}{c|}{F} & P & R & \multicolumn{1}{c|}{F} & \%75 \\ \cline{3-10}

PoseRBPF~\cite{deng2021poserbpf} & 68.73 
 & UCN+ \cite{xiang2021learning}                    & 70.0  & 81.2  & \multicolumn{1}{c|}{71.9}  & 48.3  &57.2   & \multicolumn{1}{c|}{50.3}  &   77.6   \\
 
GDRNPP~\cite{Wang_2021_GDRN} & \textbf{71.26} & MSMFormer+ \cite{lu2022mean}              & \textbf{76.1} & \textbf{83.3} & \multicolumn{1}{c|}{\textbf{76.3}} & \textbf{56.0} & \textbf{61.4} & \multicolumn{1}{c|}{\textbf{57.0}} & \textbf{80.2} \\ \hline
\end{tabular}
  }
\caption{Perception evaluation of pose estimation and unseen object instance segmentation in the real world.}
  
  \label{table:perception}
   \vspace{-5mm}
\end{table}

\begin{figure*}[h]
    \centering
    \includegraphics[width=1.6\columnwidth]{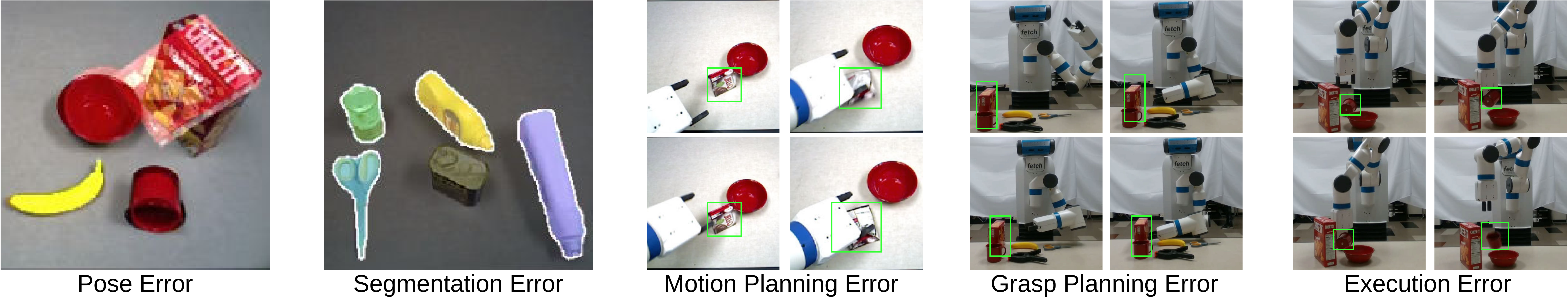}
    \vspace{-1mm}
    \caption{Example failures in our experiment pipelines. Green boxes indicate the targets for grasping.}
    \label{fig:failure-example}
    \vspace{-2mm}
\end{figure*}


\textbf{Grasping Orders.} For grasping in cluttered scenes, the order of grasping objects matters. We evaluate two orders in SceneReplica when the grasping order can be programmed: the near-to-far order based on object perception and the fixed order we defined for each scene. Intuitively, the near-to-far order is simpler, since the robot can clear up closer objects first, those which potentially occlude other objects in the scene.

\textbf{Model-based Grasping Frameworks.}  For model-based grasping, we have evaluated three 6D object pose estimation methods: PoseCNN~\cite{xiang2017posecnn} and PoseRBPF~\cite{deng2021poserbpf} and GDRNPP~\cite{Wang_2021_GDRN,liu2022gdrnpp_bop}. GraspIt!~\cite{miller2004graspit} is used for grasp planning in model-based grasping. The OMPL library~\cite{sucan2012open} in MoveIt is used to plan the trajectory of the robot arm and control to follow the planned trajectory. In cases when there is no motion plan to reach all the grasps from GraspIt!, we simply apply top-down grasping to grasp the target object.

From Table~\ref{tab:grasp_methods} we can see that the GDRNPP-based framework achieves a higher success rate in both grasping and pick-and-place. GDRNPP is the overall best method in the BOP Challenge 2022~\cite{hodan2018bop}. The accuracy of pose estimation in terms of ADD-S is presented in Table~\ref{table:perception}, which is calculated using 200 images in the real world, that is, 20 scenes $\times$ 5 grasps per scene $\times$ 2 orders. A better pose estimation model improves the success rate.

\textbf{Model-free Grasping Frameworks.} We have evaluated two unseen object segmentation methods: UCN~\cite{xiang2021learning} and MSMFormer~\cite{lu2022mean}, two grasp planning methods: 6D GraspNet~\cite{mousavian20196} and Contact-GraspNet~\cite{sundermeyer2021contact}. The OMPL library~\cite{sucan2012open} in MoveIt used for motion planning and control. Similarly, top-down grasping is used if there is no plan from the GrapsNets. From the results in Table~\ref{tab:grasp_methods}, we can see that the Contact-GraspNet-based methods achieve a higher success rate than the 6D GraspNet-based methods. Contact-GraspNet grasps usually have higher quality by modeling the contact between the gripper with the target. When comparing the two segmentation methods, MSMFormer achieves higher segmentation accuracy, as shown in Table~\ref{table:perception}. The combination of MSMFormer and Contact-GraspNet achieves the best performance among model-free grasping pipelines. MSMFormer with top-down grasping performs as a strong baseline thanks to the object segmentation network.


\textbf{End-to-end Learning-based Grasping.} End-to-end learning approaches for grasping can also be evaluated using SceneReplica. These approaches take an input image and then output a grasping action using convolutional neural networks ~\cite{mahler2017dex} (Dex-Net 2.0), \cite{morrison2018closing,kumra2020antipodal} or polices learned with reinforcement learning~\cite{levine2018learning,zeng2018learning}, \cite{kalashnikov2018qt} (QT-Opt). In particular, these methods focus mainly on top-down grasping. Until now, we have evaluated Dex-Net 2.0~\cite{mahler2017dex} on SceneReplica as shown in Table~\ref{tab:grasp_methods}, where we synthesize a top-down view depth image using the point cloud from the Fetch RGB-D camera. Its performance is inferior compared to a segmentation-based top-down grasping pipeline due to its requirement of a top-down camera and not relying on a perception module.


\textbf{Failure Analysis.}  Fig.~\ref{fig:failure-example} shows some examples of different failure types. Detailed success and failure analysis for each object and each method can be found in the supplementary material on our project website. We can see that perception errors and planning errors are dominant and that perception errors are relatively more frequent. Small and thin objects, such as scissors and markers, are more difficult to grasp.


\section{Discussion and Future Work}
\textbf{Limitations.} Our benchmark is limited by the 16 YCB objects that are used for testing. Using a more diverse set will improve the manipulation benchmarking. Second, only results with the Fetch mobile manipulator are included. Extending the benchmark to robots and external cameras would require setting up a third-person view and conducting hand-eye calibration. Once the camera is set up, we can re-render the scene reference images using the new camera pose, and these images can be used for scene recreation. 
Our experiments on SceneReplica hint at several directions to improve robot pick-and-place. First, object perception and grasp planning accuracy can be further improved. Second, one important aspect that is not considered in the model-based and model-free grasping methods is force feedback and force closure. Slippage failures can be reduced by introducing force feedback and can also improve the grasping of fragile objects.

\textbf{Future Work.} We plan to continuously evaluate more methods on SceneReplica and release these results to the community. One line of work that is missing from our current evaluation is reinforcement learning-based approaches for 6D grasping. We plan to evaluate additional end-to-end grasping approaches on SceneReplica. We hope the community can use SceneReplica to benchmark their proposed robot manipulation methods. It can also be extended to other manipulation tasks such as object rearrangement in the future.

\noindent \textbf{Acknowledgement.} This work was supported in part by the DARPA Perceptually-enabled Task Guidance (PTG) Program under contract number HR00112220005 and the Sony Research Award Program.

\label{sec:conclusion}





\vspace{-2mm}
\bibliographystyle{IEEEtran}
\bibliography{root}

\end{document}


\maketitle

\appendix

\section{Grasp and Motion Planning}
We used Graspit to compute the 6D-grasps for each object in the benchmark. The grasps were computed in an offline manner due to the time constraints imposed by Graspit. Attempting to run Graspit for each iteration in a scene would take too long time and hence, instead we decided to compute the grasps offline. We repeatedly run the grasp sampling algorithm until we obtain a desired number of grasps around the object. Finally, on the initially generated grasp set, we apply farthest point sampling on grasp translation component to ensure that the grasps are spread out across an object's surface. Fig.~\ref{fig:graspit-grasps} illustrates the generated grasps for all 16 YCB objects. In all our experiment pipelines we try top-down grasping in case motion planning fails for a given grasp. Both model-based top-down grasping and model-free top-down grasping work in a similar fashion by considering a top-down view of object's point cloud.

We compute a top-down grasp for an object by obtaining the two principal components for the point cloud's X-Y dimensions. We use the smaller component for obtaining the gripper width and it also gives the orientation to align the gripper with respect to the object. The only difference between model-based and model-free top-down grasping is that with a model-free method, we are restricted to the partial point cloud of the object. For motion planning obstacles, in model-based grasping we simply use the estimated pose with the object's known 3D mesh. In model-free methods since we don't have models, one strategy is to simply use the bounding box of the partial point cloud. However, this did not work well in practice for cluttered scenes due to large obstacle boxes for motion planning, which then inevitably fails to find a motion plan to grasp. Hence, we resort to a similar strategy as used before for top-down grasp alignment and compute an oriented bounding box using the PCA of object points in X-Y plane. 
\vspace{-2mm}
\begin{figure*}[!ht]
    \centering
    \includegraphics[width=0.8\columnwidth]{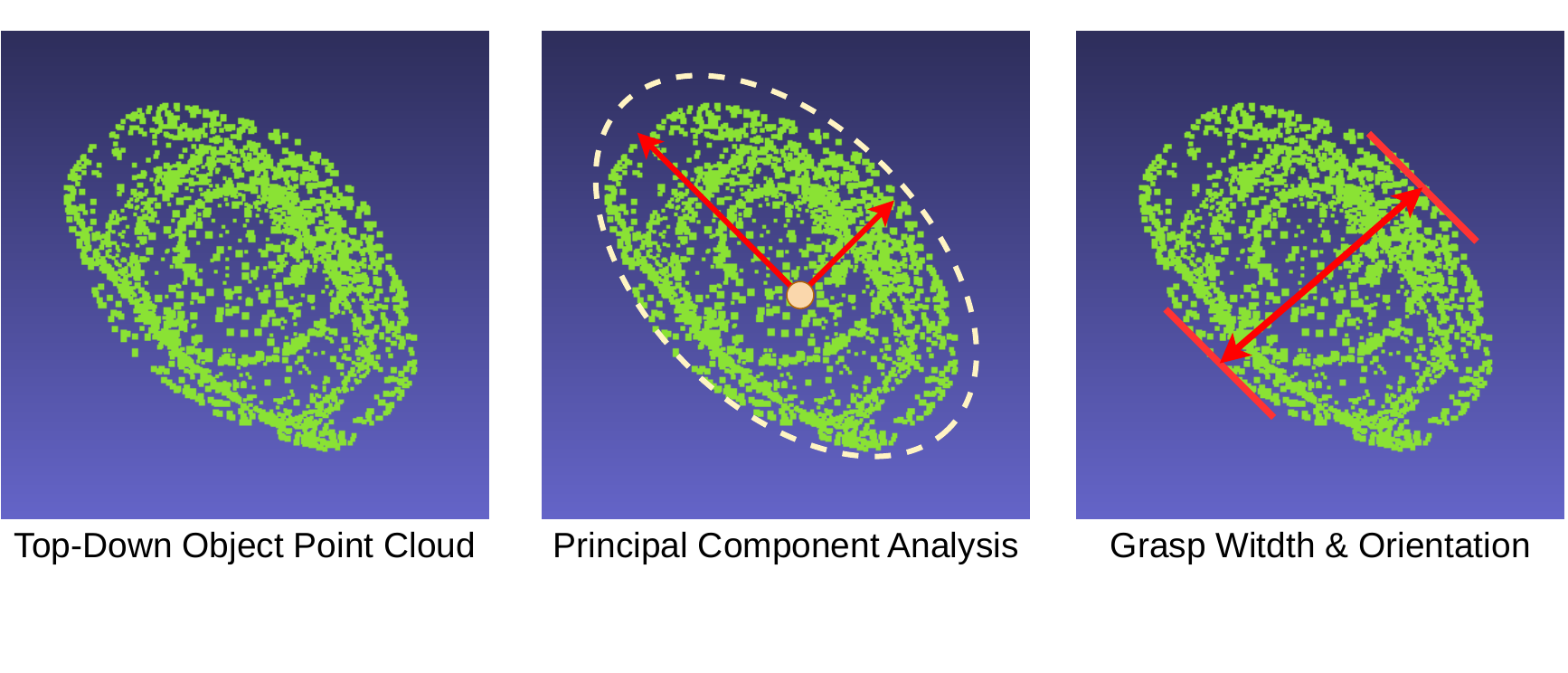}
    \vspace{-1mm}
    \caption{Top down grasp from point cloud}
    \label{fig:topdown}
    \vspace{-3mm}
\end{figure*}

\begin{figure}[!ht]
    \centering
    \begin{subfigure}{\textwidth}
        \centering
        \includegraphics[width=12cm]{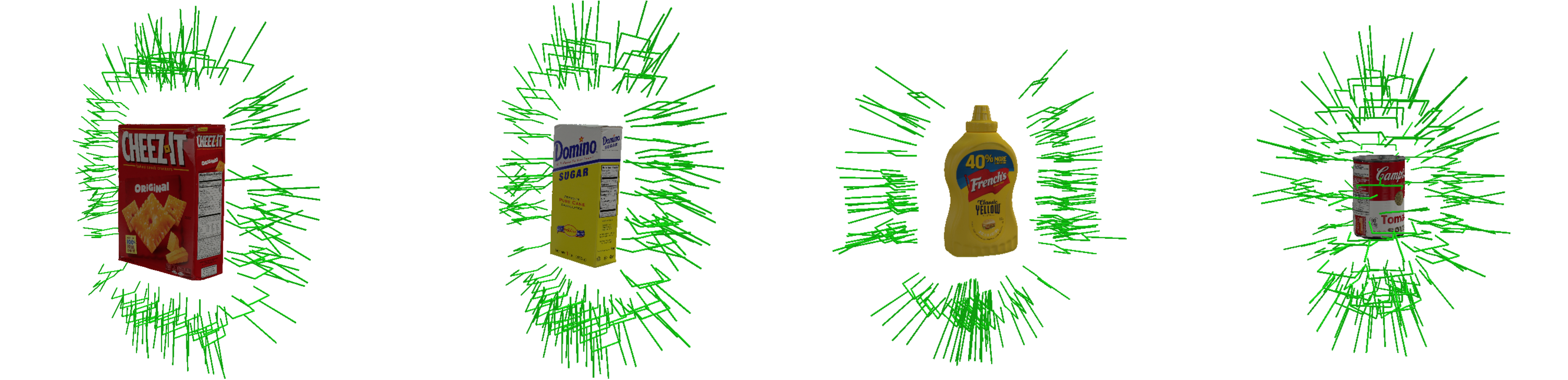}
        \vspace{1pt}
    \end{subfigure}
    
    \begin{subfigure}{\textwidth}
        \centering
        \includegraphics[width=12cm]{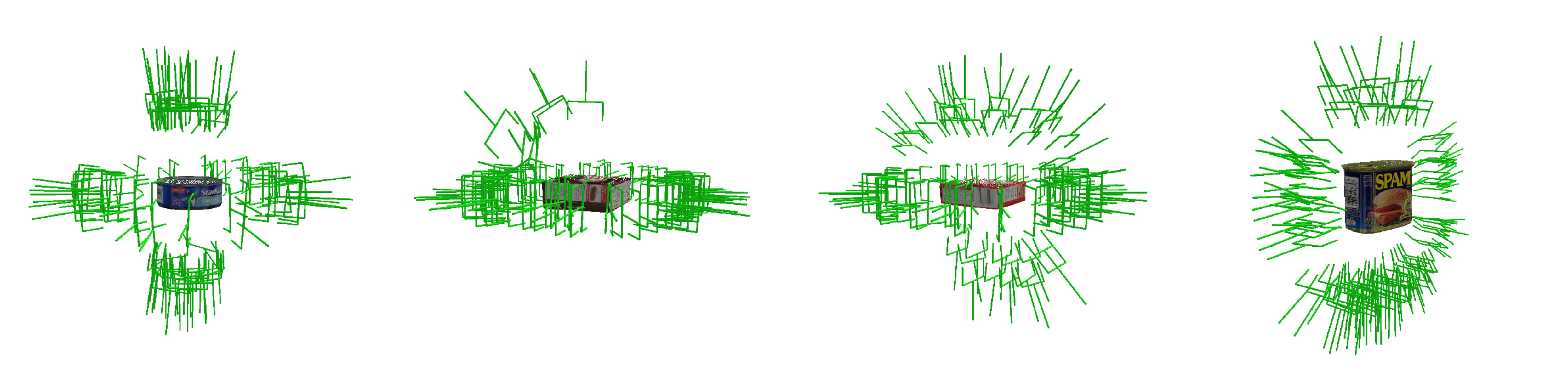}
        \vspace{1pt}
    \end{subfigure}
    
    \begin{subfigure}{\textwidth}
        \centering
        \includegraphics[width=12cm]{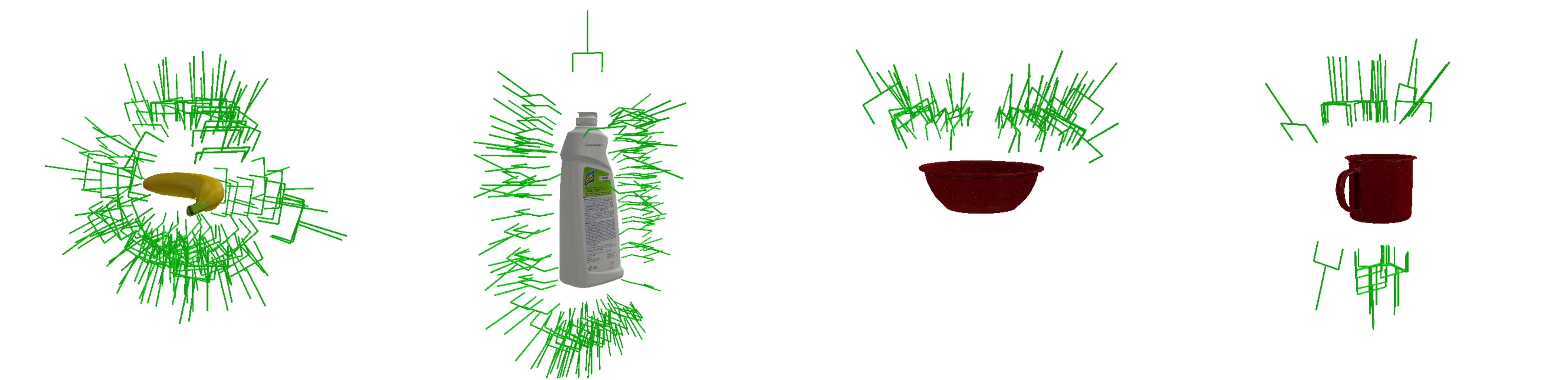}
        \vspace{1pt}
    \end{subfigure}

    \begin{subfigure}{\textwidth}
        \centering
        \includegraphics[width=12cm]{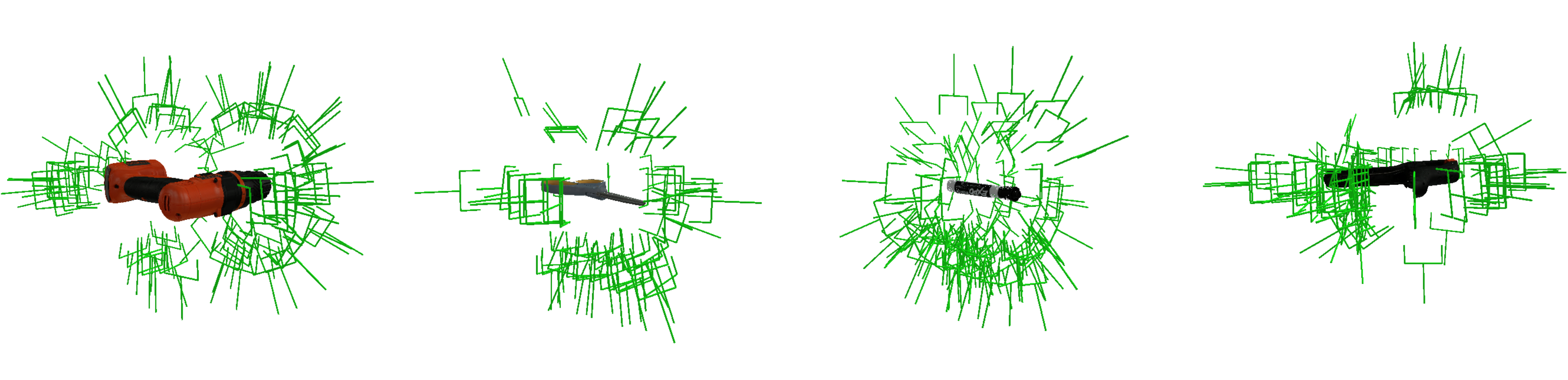}
        \vspace{1pt}
        
    \end{subfigure}
    \caption{Offline grasps generated using Graspit! for model-based grasping experiments. Pre-grasp poses shown for better visualization of underlying object.}
    \label{fig:graspit-grasps}
\end{figure}

\section{Scene Generation}
A successful and valid manipulation requires reachability of each object with respect to the robot arm. For generating the scenes in SceneReplica, we also ensure that all scenes are placed well within the robot's reachable space on the table. To find such
"safe" areas for object placement, we first discretize the table's region into a dense grid and then check for each grid location's reachability. Next, we only spawn the objects at the selected grid locations in randomly sampled stable poses, along with a random rotation along global Z axis. The random rotation along Z axis ensures that even if the same stable pose is selected for an object, the scenes for it look different. The candidate spawn locations for an object are the neighbor grid locations of already placed objects to discourage well-separated scenes, while the object spawning is made collision free by checking for collisions using their 2D bounding boxes in X-Y plane. Once we have a candidate scene with 5 collision-free spawned objects, we test motion planning to each of them using the offline grasp dataset. If any of the objects does not have a feasible motion plan, we reject the scene and move on with a new scene generation.

Such a pipeline thus ensures both diverse and feasible scenes. Finally, a set of 20 scenes is selected from hundreds of candidate scenes using criteria on (1) an object's minimum count in the set and, (2) a scoring on pose-diversity of the set. While picking a scene set from the list of generated scene, we ensure that the count distribution of objects in the set is roughly uniform. Once we have a valid set of 20 scenes, we then compute the pose-based diversity.  For a given set of 20 scenes, we first compute the count of unique poses for each object. This then also gives us a count-based probability distribution using which we compute the entropy and use that as the scoring function. The scoring functions ensures that scene-sets with the same repeated poses for objects are penalized and scene-sets with diversity are encouraged.

\section{Additional Results}

\subsection{Pick-and-place Failure Analysis}
For a pick-and-place failure observed during an experiment run, we classify  the failure into several different categories: (1) perception failure (P$_{E}$F), (2) planning failure (P$_{L}$F) and (3) execution failure (EF). Table~\ref{tab:failures-classification} describes these failures in detail. We include such detailed metrics to highlight the failure points during grasping rather than simply labeling each trial as a complete success or failure in Table~\ref{tab:grasp_methods}. The overall trend seen from Tables~\ref{tab:grasp_object_metrics_1} and \ref{tab:grasp_object_metrics_2} is that perception and planning failures dominate the error terms. Tables~\ref{tab:grasp_object_metrics_1} and \ref{tab:grasp_object_metrics_2} also show difficulties in grasping small and thin objects like the scissor, the marker, etc., due to high uncertainty in their depth and small graspable area.

\begin{table*}
\vspace{-2mm}
\centering
\resizebox{\textwidth}{!}{
\begin{tabular}{c|c|c}
\hline
Type & Phase & Description  \\ \hline

Perception failure & Pre-Grasping & Target object is not recognized, for example, no pose estimation or no segmentation mask \\
Perception failure & Pre-Grasping & Grasp not found, perception error of obstacle $>$ threshold \\
Perception failure & During-Grasping & Grasp planned, failed to grasp and lift, perception error of target $>$ threshold \\
Perception failure & During-Grasping & Grasp planned,  hit obstacle, perception error of obstacle $>$ threshold \\ \hline
Planning failure & Pre-Grasping & No perception failure, but no plan found to grasp target \\ 
Planning failure & During-Grasping & Grasp planned, no perception failure, failed to grasp and lift or hit obstacle \\ \hline
Execution failure & Post-Grasping & Failed to place object after grasping and lifting \\
\hline
\end{tabular}
}
\caption{Different types of failures in our pick-and-place experimental analysis.}
\label{tab:failures-classification}
\end{table*}

\begin{table*}
\vspace{-1mm}
\centering
\resizebox{\textwidth}{!}{
\begin{tabular}{c|c|c|c|c|ccc}
\hline
Method \# & Perception & Grasp Planning & Motion Planning & Control & Ordering & Pick-and-Place Success & Grasping Success  \\ \hline

\multicolumn{8}{c}{Model-based Grasping} \\ \hline

1 & PoseRBPF~\cite{deng2021poserbpf} & GraspIt!~\cite{miller2004graspit} + Top-down & OMPL~\cite{sucan2012open} & 
 MoveIt  & Near-to-far & 58 / 100 & 64 / 100\\

1 & PoseRBPF~\cite{deng2021poserbpf} & GraspIt!~\cite{miller2004graspit} + Top-down & OMPL~\cite{sucan2012open} & MoveIt  & Fixed & 59 / 100 & 59 / 100\\

2 & PoseCNN~\cite{xiang2017posecnn} & GraspIt!~\cite{miller2004graspit} + Top-down & OMPL~\cite{sucan2012open} & 
 MoveIt  & Near-to-far & 47 / 100 & 48 / 100\\

2 & PoseCNN~\cite{xiang2017posecnn}  & GraspIt!~\cite{miller2004graspit} + Top-down & OMPL~\cite{sucan2012open} & MoveIt  & Fixed & 40 / 100 & 45 / 100 \\

3 & GDRNPP~\cite{Wang_2021_GDRN,liu2022gdrnpp_bop} & GraspIt!~\cite{miller2004graspit} + Top-down & OMPL~\cite{sucan2012open} & 
 MoveIt  & Near-to-far & \textbf{66 / 100} & 69 / 100\\

3 & GDRNPP~\cite{Wang_2021_GDRN,liu2022gdrnpp_bop} & GraspIt!~\cite{miller2004graspit} + Top-down & OMPL~\cite{sucan2012open} & MoveIt  & Fixed & 62 / 100 & 64 / 100\\ \hline

\multicolumn{8}{c}{Model-free Grasping} \\ \hline

4 & UCN~\cite{xiang2021learning} & GraspNet~\cite{mousavian20196} + Top-down & OMPL~\cite{sucan2012open} & 
 MoveIt  & Near-to-far & 43 / 100  & 46 / 100\\

 4 & UCN~\cite{xiang2021learning} & GraspNet~\cite{mousavian20196} + Top-down & OMPL~\cite{sucan2012open} & 
 MoveIt  & Fixed & 37 / 100 & 40 / 100 \\

 5 & UCN~\cite{xiang2021learning} & Contact-graspnet~\cite{sundermeyer2021contact} + Top-down & OMPL~\cite{sucan2012open} & 
 MoveIt & Near-to-far & 60 / 100  & 63 / 100\\

  5 & UCN~\cite{xiang2021learning} & Contact-graspnet~\cite{sundermeyer2021contact} + Top-down & OMPL~\cite{sucan2012open} & 
 MoveIt  & Fixed & 60 / 100 & 64 / 100 \\

 6 & MSMFormer~\cite{lu2022mean} & GraspNet~\cite{mousavian20196} + Top-down & OMPL~\cite{sucan2012open} & 
 MoveIt  & Near-to-far & 38 / 100 & 41 / 100 \\

 6 & MSMFormer~\cite{lu2022mean} & GraspNet~\cite{mousavian20196} + Top-down & OMPL~\cite{sucan2012open} & 
 MoveIt  & Fixed & 36 / 100 & 41 / 100 \\

 7 &  MSMFormer~\cite{lu2022mean} &  Contact-graspnet~\cite{sundermeyer2021contact} + Top-down & OMPL~\cite{sucan2012open} & 
 MoveIt  & Near-to-far & 57 / 100  & 65 / 100\\ 

 7 &  MSMFormer~\cite{lu2022mean} &  Contact-graspnet~\cite{sundermeyer2021contact} + Top-down & OMPL~\cite{sucan2012open} & 
 MoveIt  & Fixed & 61 / 100  & \textbf{70 / 100}\\

8 & MSMFormer~\cite{lu2022mean} & Top-down & OMPL~\cite{sucan2012open} & 
 MoveIt  & Fixed & 56 / 100  & 59 / 100\\
\hline
\multicolumn{8}{c}{End-to-end Learning-based Grasping} \\ \hline

9 & \multicolumn{2}{c|}{Dex-Net 2.0~\cite{mahler2017dex} (Top-Down Grasping)} & OMPL~\cite{sucan2012open} &  MoveIt & Algorithmic &  43 /100   &  51 / 100 \\

\hline
\end{tabular}
}
\vspace{-2mm}
\caption{Different grasping frameworks evaluated on SceneReplica using a Fetch mobile manipulator.}
\label{tab:grasp_methods}
\vspace{-3mm}
\end{table*}

\begin{table*}
\resizebox{\textwidth}{!}
{

\begin{tabular}{l|c|cccc|cccc|cccc|cccc} 
\hline
\multirow{2}{*}{Object} & \multirow{2}{*}{Count} & \multicolumn{4}{c|}{Method 1} &  \multicolumn{4}{c|}{Method 2}  & \multicolumn{4}{c|}{Method 3} &  \multicolumn{4}{c}{Method 4}  \\ \cline{3-18}

&  & S & P$_{E}$F & P$_{L}$F & EF & S & P$_{E}$F & P$_{L}$F & EF & S & P$_{E}$F & P$_{L}$F & EF & S & P$_{E}$F & P$_{L}$F & EF  \\ \hline

\multicolumn{18}{c}{Order: Near-to-Far} \\ \hline

003 cracker box & 6 & 5 & - & 1 & - 
                    & 1 & 4 & 1 & - 
                    & 3 & 2 & 1 & -
                    & 2 & 2 & 2 & - 
                    
                    \\
004 sugar box & 5 & 5 & - & - & - 
                    & 1 & 4 & - & -   
                    & 5 & - & - & - 
                    & 2 & 2 & 1 & - 
                    
                    \\
005 tomato soup can & 7 & 6 & 1 & - & - 
                        & 6 & 1 & - & -  
                        &  5 & 1 & - & 1  
                        & 3 & - & 4 & - 
                        
                        \\
006 mustard bottle & 7 & 6 & 1 & - & - 
                        & 3 & 2 & 2 & -
                        & 7 & - & - & -
                        & 2 & 1 & 4 & - 
                        
                        \\
    007 tuna fish can & 6 & 1 & 1 & 4 & - 
                        & 3 & 2 & 1 & -
                        & 1 & 5 & - & -
                        & 6 & - & - & -
                        
                        \\
008 pudding box & 5 & 5 & - & - & - 
                    & 4 & 1 & - & -
                    &  5 & - & - & - 
                    & 3 & - & 2 & -
                    
                    \\
009 gelatin box & 7 & 3 & 4 & - & - 
                    & 2 & 4 & 1 & -
                    & 6 & - & 1 & -
                    & 3 & 2 & 1 & 1
                    
                    \\
010 potted meat can & 7 & 6 & 1 & - & - 
                    & 3 & 2 & 2 & -
                    & 7 & - & - & -
                    & 4 & 1 & 2 & -
                    
                    \\
011 banana & 7 & 4 & - & 2 & 1 
                    & 2 & 4 & 1 & -
                    & 6 & - & 1 & -
                    & 1 & - & 6 & -
                    
                    \\
021 bleach cleanser & 5 & 3 & - & - & 2 
                    & 1 & 1 & 1 & 2
                    & 3 & 1 & - & 1
                    & 2 & 1 & 1 & 1
                    
                    \\
024 bowl & 7 & 2 & 4 & 1 & - 
                    & 5 & 2 & - & -
                    & 2 & 4 & 1 & - 
                    & 5 & - & 2 & -
                    
                    \\
025 mug & 5 & 2 & 1 & - & 2 
                    & 3 & 2 & - & -
                    & 4 & - & 1 & -
                    & 3 & 1 & 1 & -
                    
                    \\
037 scissors & 7 & 1 & 2 & 4 & - 
                    & 1 & 3 & 3 & -
                    & 4 & 3 & - & -
                    & 1 & - & 5 & 1
                    
                    \\
035 power drill & 7 & 2 & 1 & 3 & 1 
                    & 3 & 2 & 2 & -
                    & 1 & 3 & 2 & 1
                    & - & 7 & - & -
                   
                    \\
040 large marker & 6 & 1 & 4 & 1 & - 
                    & 4 & 1 & 1 & - 
                    & 2 & 4 & - & -
                    & 3 & 1 & 2 & -
                    
                    \\
052 extra large clamp & 6 & 6 & - & - & - 
                    & 5 & 1 & - & -
                    & 5 & 1 & - & - 
                    & 3 & - & 3 & -
                   
                    \\ \hline
ALL & 100 & 58 & 20 & 16 & 6 &
                    47 & 36 & 15 & 2 & 66 & 24 & 7 & 3 & 43 & 18 & 36 & 3 \\

\hline

\multicolumn{18}{c}{Order: Randomly-Fixed} \\ \hline

003 cracker box & 6 & 5 & 1 & - & -  
                    & 2 & 4 & - & -  
                    & 5 & 1 & - & -  
                    & 2 & 3 & 1 & -    
                   
                    \\
004 sugar box & 5 & 4 & 1 & - & -  
                  & 1 & 3 & 1 & -
                  & 5 & - & - & -  
                  & 3 & - & 2 & -   
                  
                  \\
005 tomato soup can & 7 & 7 & - & - & -  
                        & 6 & 1  & -  &  - 
                        & 6 & 1  & -  &  - 
                        & 2 & 1 & 4 & -   
                        
                        \\
006 mustard bottle & 7 & 7 & - & - & -  
                       & 2 & 4 & 1 & -  
                       & 5 & 1  & 1  &  - 
                       & 3 & 2 & 2 & - 
                       
                       \\
007 tuna fish can & 6 & 2 & 4 & - & -  
                      & 1 & 4 & - & 1 
                      & 2 & 3 & 1 & -
                      & 4 & 1 & - & 1 
                     
                      \\
008 pudding box & 5 & 4 & - & 1 & -  
                    & 3 & 1 & 1 & - 
                    & 3 & 2 & - & -
                    & 3 & - & 2 & - 
                    
                    \\
009 gelatin box & 7 
                    & 3  &  4 &  - & -  
                    & 4 & - & 2 & 1 
                    & 5 & 2 & - & -
                    & 7 & - & - & -
                    \\
010 potted meat can & 7 & 7 & - & - & -  
                    & 5  &  1 &  1 &  - 
                    & 7 & - & - & -
                    & 1 & 3 & 2 & 1
                    
                    \\
011 banana & 7      & 2 & 1 & 4 & -  
                    &  2 &  4 &  1 & -  
                    & 3 & 2 & 2 & -
                    & 5 & - & 2 & -
                   
                    \\
021 bleach cleanser & 5 & 3 & 2 & - & -  
                    & 1 & 3  & 1  & -  
                    & 4 & - & - & 1
                    & 1 & 1 & 2 & 1
                    
                    \\
024 bowl & 7        & 4 & 3 & - & -  
                    & 1  & 4  & 2  & -  
                    & 2 & 4 & 1 & -
                    & 3 & 2 & 2 & -
                    
                    \\
025 mug & 5         & 5 & - & - & -  
                    & 3  & 2  & -  & -  
                    & 3 & - & 1 & 1
                    & 4 & - & 1 & -
                    
                    \\
037 scissors & 7    & - & 1 & 6 & -  
                    & - & 2  & 5  & - 
                    & 4 & 3 & - & -
                    & - & 3 & 4 & -
                    
                    \\
035 power drill & 7 & 2 & 1 & 4 & -  
                    & 2  & 2  & 3  &  - 
                    & - & 6 & 1 & -
                    & - & 7 & - & -
                    
                    \\
040 large marker & 6 & - & 5 & 1 & -  
                    & 3  &  3 &  - &  - 
                    & 2 & 3 & 1 & -
                    & - & 1 & 5 & -
                   
                    \\
052 extra large clamp & 6   & 6 & - & - & -  
                            & 5  &  - &  1 & -  
                            & 6 & - & - & -  
                            & 2 & - & 4 & -
                           
                            \\
                    \hline
ALL & 100 & 59 & 25 & 16 & - & 40 & 42 & 17 & 1 & 62 & 28 & 8 & 2 & 37 & 24 & 35 & 4   \\ \hline

\multicolumn{18}{c}{} \\
\multicolumn{18}{c}{} \\ \hline

\multirow{2}{*}{Object} & \multirow{2}{*}{Count} & \multicolumn{4}{c|}{Method 5} &  \multicolumn{4}{c|}{Method 6}  & \multicolumn{4}{c|}{Method 7} &  \multicolumn{4}{c}{Method 8} \\ \cline{3-18}

&  & S & P$_{E}$F & P$_{L}$F & EF & S & P$_{E}$F & P$_{L}$F & EF & S & P$_{E}$F & P$_{L}$F & EF & S & P$_{E}$F & P$_{L}$F & EF  \\ \hline

\multicolumn{18}{c}{Order: Near-to-Far} \\ \hline

003 cracker box & 6 & 3 & 1 & 2 & - & 2 & 3 & 1 & - 
                & 4 & 1 & - & 1
                
                &  \textcolor{red}{\ding{55}} &  \textcolor{red}{\ding{55}} & \textcolor{red}{\ding{55}}  & \textcolor{red}{\ding{55}} \\
004 sugar box & 5 & 5 & - & - & -
                    & 3 & 1 & 1 & -
                    & 5 & - & - & -
                    
                    &  \textcolor{red}{\ding{55}} &  \textcolor{red}{\ding{55}} & \textcolor{red}{\ding{55}}  & \textcolor{red}{\ding{55}} \\
 005 tomato soup can & 7 & 6 & - & 1 & - 
                        & 2 & - & 5 & -
                        & 2 & 2 & 3 & -
                        
                        &  \textcolor{red}{\ding{55}} &  \textcolor{red}{\ding{55}} & \textcolor{red}{\ding{55}}  & \textcolor{red}{\ding{55}} \\
006 mustard bottle & 7 & 5 & 1 & 1 & -
                        & 1 & - & 5 & 1
                        & 6 & - & 1 & -
                        
                        &  \textcolor{red}{\ding{55}} &  \textcolor{red}{\ding{55}} & \textcolor{red}{\ding{55}}  & \textcolor{red}{\ding{55}} \\
007 tuna fish can & 6 & 5 & 1 & - & -
                        & 5 & - & 1 & -
                        & 5 & 1 & - & -
                        
                        &  \textcolor{red}{\ding{55}} &  \textcolor{red}{\ding{55}} & \textcolor{red}{\ding{55}}  & \textcolor{red}{\ding{55}} \\
008 pudding box & 5 & 4 & - & 1 & -
                    & 4 & - & 1 & -
                    & 4 & 1 & - & -
                    
                    &  \textcolor{red}{\ding{55}} &  \textcolor{red}{\ding{55}} & \textcolor{red}{\ding{55}}  & \textcolor{red}{\ding{55}} \\
009 gelatin box & 7 & 7 & - & - & -
                    & 4 & - & 3 & -
                    & 6 & - & 1 & -
                    
                    &  \textcolor{red}{\ding{55}} &  \textcolor{red}{\ding{55}} & \textcolor{red}{\ding{55}}  & \textcolor{red}{\ding{55}} \\
010 potted meat can & 7 & 3 & 2 & 1 & 1
                    & 1 & - & 6 & -
                    & 5 & 2 & - & -
                    
                    &  \textcolor{red}{\ding{55}} &  \textcolor{red}{\ding{55}} & \textcolor{red}{\ding{55}}  & \textcolor{red}{\ding{55}} \\
011 banana & 7 & 2 & - & 5 & -
                    & 5 & - & 2 & -
                    & 6 & - & - & 1
                    
                    &  \textcolor{red}{\ding{55}} &  \textcolor{red}{\ding{55}} & \textcolor{red}{\ding{55}}  & \textcolor{red}{\ding{55}} \\
021 bleach cleanser & 5 & 2 & - & 2 & 1
                    & - & 1 & 3 & 1
                    & - & 1 & 2 & 2
                    
                    &  \textcolor{red}{\ding{55}} &  \textcolor{red}{\ding{55}} & \textcolor{red}{\ding{55}}  & \textcolor{red}{\ding{55}} \\
024 bowl & 7 & 7 & - & - & -
                    & 5 & - & 1 & 1
                    & 6 & - & - & 1
                    
                    &  \textcolor{red}{\ding{55}} &  \textcolor{red}{\ding{55}} & \textcolor{red}{\ding{55}}  & \textcolor{red}{\ding{55}} \\
025 mug & 5 & 1 & 1 & 3 & -
                    & 2 & - & 3 & -
                    & 2 & - & 2 & 1
                    
                    &  \textcolor{red}{\ding{55}} &  \textcolor{red}{\ding{55}} & \textcolor{red}{\ding{55}}  & \textcolor{red}{\ding{55}} \\
037 scissors & 7 & 3 & 2 & 2 & -
                    & - & 2 & 5 & -
                    & - & 2 & 2 & 3
                    
                    &  \textcolor{red}{\ding{55}} &  \textcolor{red}{\ding{55}} & \textcolor{red}{\ding{55}}  & \textcolor{red}{\ding{55}} \\
035 power drill  & 7 & 2 & 4 & - & 1
                    & 1 & 3 & 3 & -
                    & 3 & 3 & - & 1
                    
                    &  \textcolor{red}{\ding{55}} &  \textcolor{red}{\ding{55}} & \textcolor{red}{\ding{55}}  & \textcolor{red}{\ding{55}} \\
040 large marker & 6 & 1 & 2 & 3 & -
                    & 3 & - & 3 & -
                    & 1 & 2 & 2 & 1
                    
                    &  \textcolor{red}{\ding{55}} &  \textcolor{red}{\ding{55}} & \textcolor{red}{\ding{55}}  & \textcolor{red}{\ding{55}} \\
052 extra large clamp & 6  & 4 & 1 & 1 & -
                    & - & 1 & 5 & -
                    & 2 & 1 & 2 & 1
                    
                    &  \textcolor{red}{\ding{55}} &  \textcolor{red}{\ding{55}} & \textcolor{red}{\ding{55}}  & \textcolor{red}{\ding{55}} \\
\hline
ALL & 100  &  60 & 15 & 22 & 3 & 38 & 11 & 48 & 3 & 57 & 16 & 15 & 12 & \textcolor{red}{\ding{55}} &  \textcolor{red}{\ding{55}} & \textcolor{red}{\ding{55}}  & \textcolor{red}{\ding{55}} \\
\hline

\multicolumn{18}{c}{Order: Randomly-Fixed} \\ \hline

003 cracker box  & 6 & 4 & 1 & 1 & - 
                    & 3 & 2 & 1 & -
                    & 4 & 1 & 1 & -
                    
                    & 2  & 3 & 1 & - \\
004 sugar box & 5 & 3 & - & 2 & -
                  & 3 & 1 & 1 & -
                  & 5 & - & - & -
                  
                    &  5 & - & - & -  \\
 005 tomato soup can & 7 & 6 & 1 & - & - 
                        & 1 & - & 5 & 1
                        & 7 & - & - & -
                        
                    &  4 & - & 1 & 2 \\
006 mustard bottle & 7 & 1 & 1 & 3 & 1 
                       & - & - & 6 & 1
                       & 3 & - & 1 & 3
                       
                    &  6 & - & 1 & - \\
007 tuna fish can & 6  & 4 & 1 & 1 & -
                      & 2 & 1 & 3 & -
                      & 4 & 1 & - & 1
                      
                    & 5  & - & 1 & - \\
008 pudding box & 5 & 4 & 1 & - & -
                    & 4 & - & 1 & -
                    & 4 & - & - & 1
                    
                    &  5 & - & - & - \\
009 gelatin box & 7 & 1 & 6 & - & -  
                    & 4 & 1 & 2 & -
                    & 6 & - & 1 & -
                    
                    & 7  & - & - & -  \\
010 potted meat can & 7 & 6 & 1 & - & -
                    & 4 & 1 & 2 & -
                    & 4 & - & 2 & 2
                    
                    &  4 & 1 & 2  & - \\
011 banana & 7  & 1 & - & 6 & -
                    & 3 & 1 & 2 & 1
                    & 5 & 1 & 1 & -
                    
                    &  3 & 1 &  3&  - \\
021 bleach cleanser & 5 & 1 & 1 & 2 & 1
                    & 1 & 1 & 3 & -
                    & 1 & - & 2 & 2
                    
                    &  3 & 1 & 1 & -  \\
024 bowl & 7         & 7 & - & - & -
                    & 4 & - & 2 & 1
                    & 7 & - & - & -
                    
                    & 7 & - & - & - \\
025 mug & 5 & 4 & - & - & 1
                    & 2 & 1 & 2 & -
                    & 2 & - & 3 & -
                    
                    & 1 & 1 & 3 & -\\
037 scissors & 7 & 2 & 2 & 3 & -
                    & 2 & 2 & 3 & -
                    & 2 & 4 & 1 & -
                    
                    &  1 & 4 &  2 & - \\
035 power drill  & 7 & 3 & 4 & - & -
                    & - & 3 & 3 & 1
                    & 4 & 1 & 1 & 1
                    
                    &  1 & 4 & 2 & - \\
040 large marker & 6  & 3 & 2 & 1 & -
                    & 2 & 1 & 3 & -
                    & 2 & 4 & - & -
                    
                    &  2 & 2 & 1 & 1 \\
052 extra large clamp & 6  & 3 & 2 & 1 & -
                            & 1 & - & 5 & -
                            & 2 & 2 & 2 & - 
                        
                    &  - & 5 & 1 & - \\
                    \hline
ALL & 100 &  60 & 17 & 20 & 3 & 36 & 15 & 44 & 5 & 61 & 14 & 15 & 10 & 56 & 22 & 19 & 3\\
\hline

\end{tabular}
}
\caption{ Statistics of our grasping experiments for each YCB object (Methods\hspace{1pt}1-8). S: \#pick-and-place success, P$_{E}$F: \#perception failure, P$_L$F: \#planning failure, EF: \#execution failure}.
\label{tab:grasp_object_metrics_1}
\vspace{-4mm}
\end{table*}

\begin{table*}
\centering
\resizebox{0.55\textwidth}{!}
{
\begin{tabular}{l|c|cccc} 
\hline
\multirow{2}{*}{Object} & \multirow{2}{*}{Count} & \multicolumn{4}{c}{Method 9}  \\ \cline{3-6}
&  & S & P$_{E}$F & P$_{L}$F & EF \\ \hline
\multicolumn{6}{c}{Order: Algorithmic} \\ \hline
003 cracker box  & 6 & -  & - & 5 & 1
                   \\
004 sugar box & 5  &  4 & - & 1 & - 
                  \\
 005 tomato soup can & 7 &  2 & - & 4 & 1
                     \\
006 mustard bottle & 7 &  2 & - & 4 & 1
                     \\
007 tuna fish can & 6  & -  & - & 6 & -
                     \\
008 pudding box & 5 &  4 & - & 1 & -
                    \\
009 gelatin box & 7 & 6  & - & 1 & -  
                    \\
010 potted meat can & 7 &  5 & - & 2  & - 
                    \\
011 banana & 7  &  6 & - &  1&  -
                    \\
021 bleach cleanser & 5 &  - & - & 4 & 1 
                    \\
024 bowl & 7         &  6 & - & - & 1
                    \\
025 mug & 5 & 2 & - & 3 & -
                    \\
037 scissors & 7 &  - & - &  7 & -
                    \\
035 power drill  & 7 &  - & - & 6 & 1
                    \\
040 large marker & 6  &  3 & - & 2 & 1
                    \\
052 extra large clamp & 6  &  3 & - & 2 & 1
                    \\ \hline
ALL & 100 &  43 & - & 49 & 8 \\
                    \hline
\end{tabular}
}
\caption{Statistics of our grasping experiments for each YCB object (Method\#9). S: \#pick-and-place success, P$_{E}$F: \#perception failure, P$_L$F: \#planning failure, EF: \#execution failure}
\label{tab:grasp_object_metrics_2}
\vspace{-4mm}
\end{table*}

\subsection{Pose Estimation Validation}
\begin{figure}[!ht]
    \centering
    \includegraphics[width=0.98\linewidth]{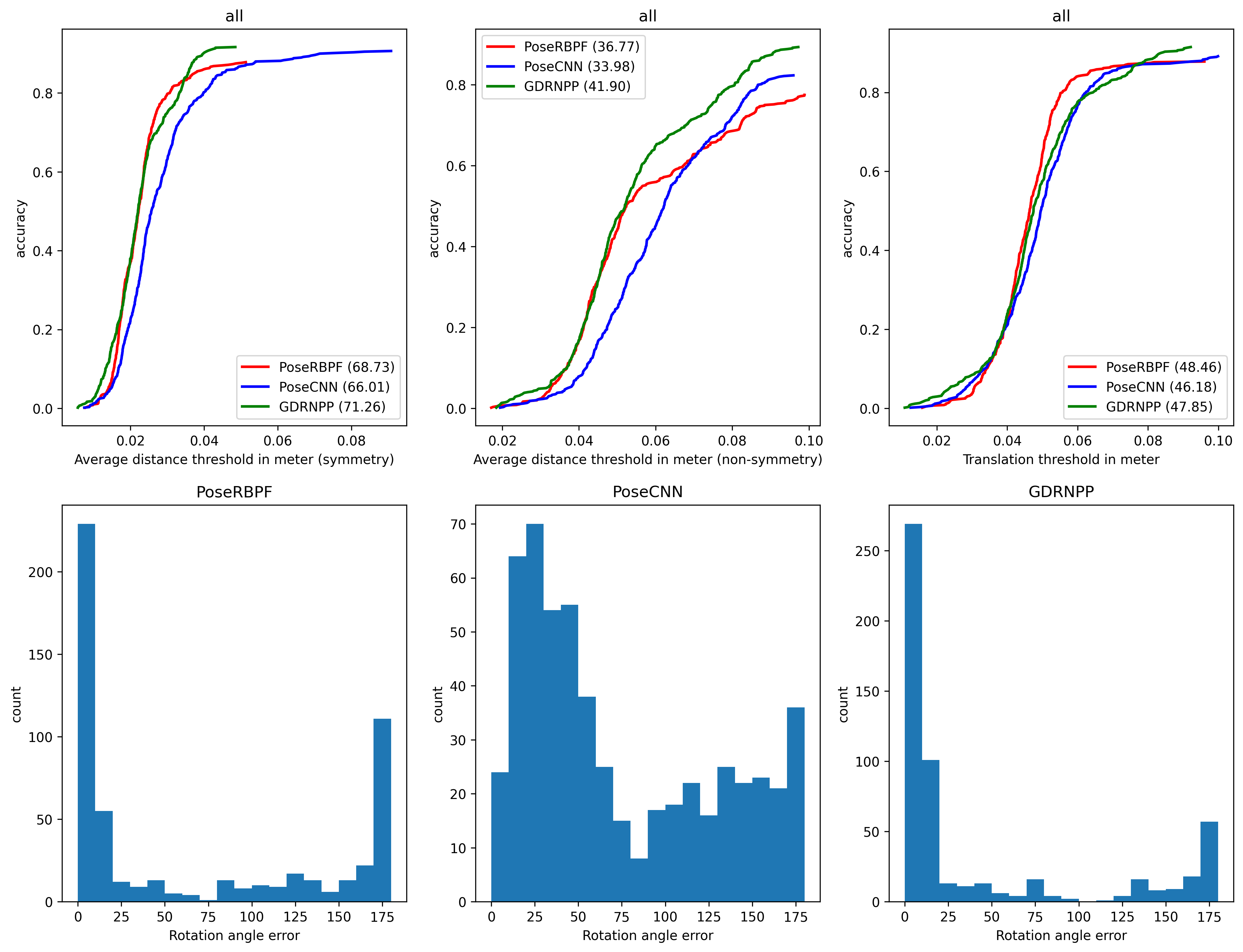}
    \caption{Visualization of pose estimation validation}
    \label{fig:pose_results}
\end{figure}

We have also visualized the results for the validation on pose estimation methods where we plot an ADD threshold-accuracy curve and a frequency histogram for pose rotation angle error as shown in Figure~\ref{fig:pose_results}. In the threshold-accuracy curves, we tweak the tolerance of translation and rotation difference with ground truth for classifying a pose prediction as successful or unsuccessful. As we can see, the latest GDRNPP~\cite{Wang_2021_GDRN,liu2022gdrnpp_bop} method outperforms other methods in terms of pose estimation which also translates to a better result in pick-and-place grasping as shown in Table~\ref{tab:grasp_object_metrics_1}.



    


\section{Scene Replication for Different Robot Platform}
\label{sec:diff-robot-platform}
\subsection{Robot Reachable Space Verification}
We show an example about extending the scene reachability for a different platform with a Franka-Panda arm. The arm is loaded in the simulation environment with just slight changes in height owing to the difference in morphology (we assume researchers have access to height adjustable table as noted in the paper). 
Then using a similar procedure as that for Fetch robot, we spawn small cubes in dense grid locations on the table. A valid motion plan to each cube is checked and cubes with no plan are removed. This gives us an idea about the reachable space of the new robot as seen in Fig.2 of the paper. The scenes can then be spawned inside this reachable space and reference images are generated given the choice of custom camera parameters as described in Section~\ref{sec:ref-image-generation}.

\subsection{Reference Image Generation with Different Camera}
\label{sec:ref-image-generation}
Here we demonstrate a replication procedure for the reference scene images under different camera settings. With a different robot platform, the reference scene images can be regenerated in simulation given the new camera's parameters. Using the outlined procedure the reference images can be generated even if the camera is separate from the robot platform. 

\begin{itemize}
    \item Spawn the objects within the reachable space of new robot as shown in Section~\ref{sec:diff-robot-platform}.
    
    \item Adjust the robot's camera by changing the pose so that all objects are visible.

    \item Render the images in simulation using the camera and use them as reference while setting up the scene in real-world. 
\end{itemize}

For an example, refer to Table~\ref{tab:scenes} which contains reference images for two scenes generated using the same ground truth poses of the objects, but under two different camera settings. Note that the real-world camera parameters should match the one used in simulation for accurate scene replication. 

\begin{table}[!h]
    \centering
    \begin{tabular}{|c|c|c|}
        \hline
        \multicolumn{3}{|c|}{Generated Scenes} \\ \hline
        Fetch & Camera placement I & Camera placement II \\
        \includegraphics[width=0.3\linewidth]{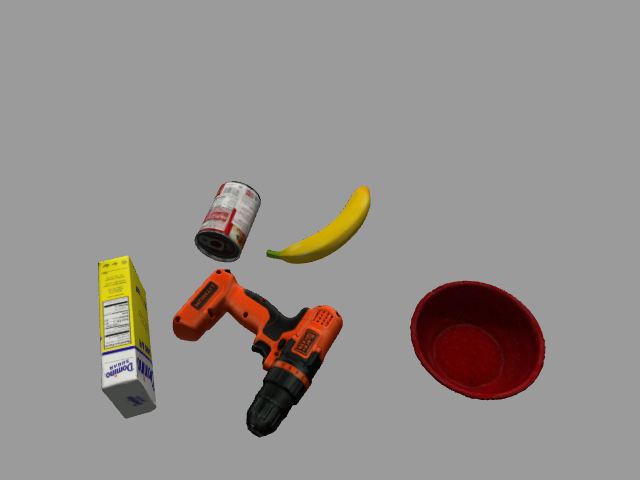} & \includegraphics[width=0.3\linewidth]{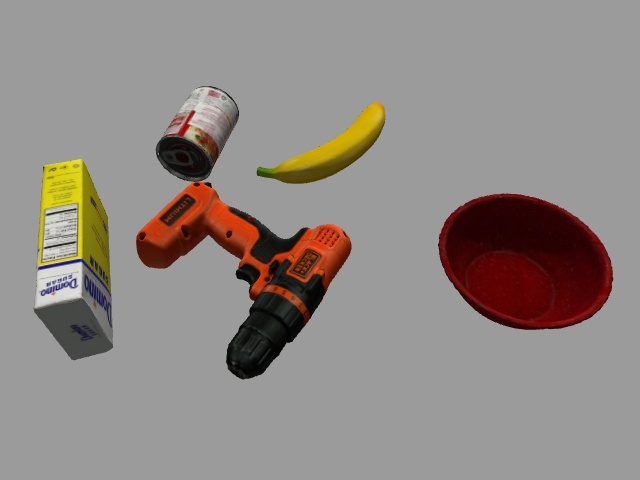} & \includegraphics[width=0.3\linewidth]{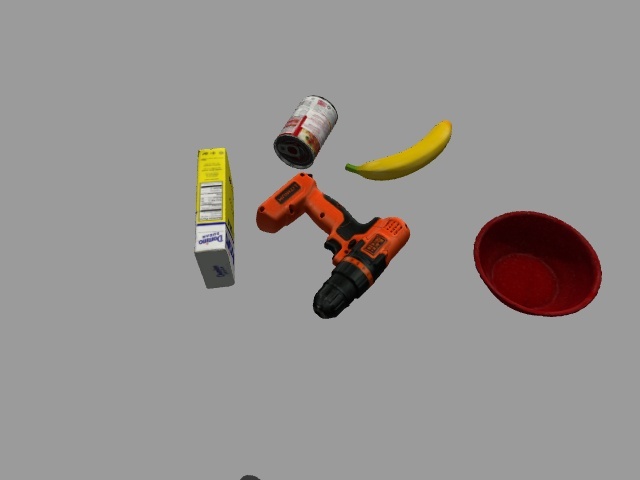} \\ 
        \includegraphics[width=0.3\linewidth]{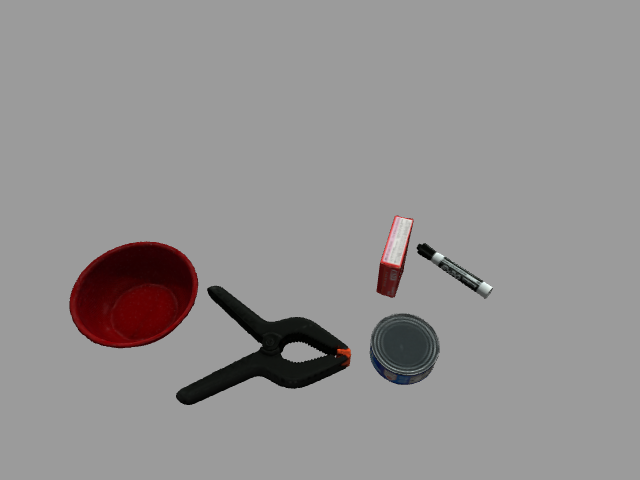} & \includegraphics[width=0.3\linewidth]{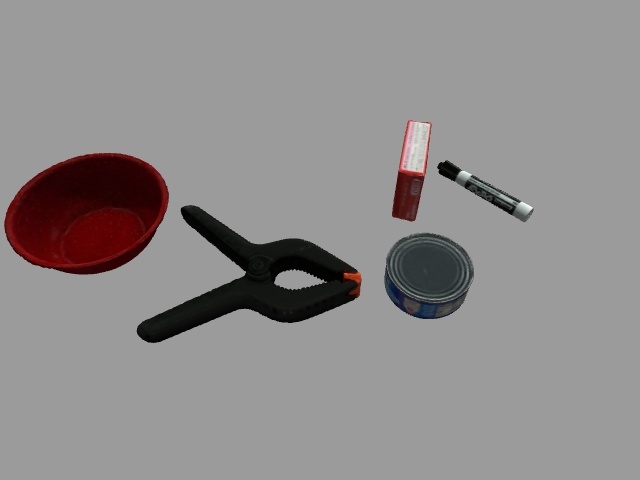} & \includegraphics[width=0.3\linewidth]{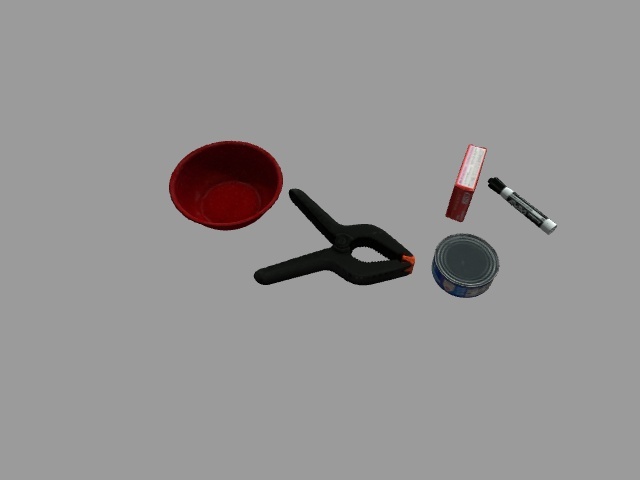} \\ \hline
    \end{tabular}
    \caption{Reference images with two different camera settings}
    \label{tab:scenes}
\end{table}

\bibliographystyle{IEEEtran}
\bibliography{root}

%% file: includes/title.tex
\LARGE \bf \textsc{SceneReplica}: Benchmarking Real-World Robot Manipulation by Creating Replicable Scenes

%% file: includes/authors.tex
Ninad Khargonkar$^{*}$, Sai Haneesh Allu$^{*}$, Yangxiao Lu, Jishnu Jaykumar P \\ Balakrishnan Prabhakaran, Yu Xiang
\thanks{$^{*}$ Equal contribution, the authors are with the Department of Computer Science, University of Texas at Dallas, Richardson, TX 75080, USA {\tt\small \{firstname.lastname\}@utdallas.edu}}%

%% file: includes/abstract.tex
We present a new reproducible benchmark for evaluating robot manipulation in the real world, specifically focusing on a pick-and-place task. Our benchmark uses the YCB object set, a commonly used dataset in the robotics community, to ensure that our results are comparable to other studies. Additionally, the benchmark is designed to be easily reproducible in the real world, making it accessible to researchers and practitioners. We also provide our experimental results and analyzes for model-based and model-free 6D robotic grasping on the benchmark, where representative algorithms are evaluated for object perception, grasping planning, and motion planning. We believe that our benchmark will be a valuable tool for advancing the field of robot manipulation. By providing a standardized evaluation framework, researchers can more easily compare different techniques and algorithms, leading to faster progress in developing robot manipulation methods. \footnote{Appendix, code and videos for the project are available at \\ \url{https://irvlutd.github.io/SceneReplica}}
